\documentclass[%
 aip,
 amsmath,amssymb,
preprint,%
]{revtex4-2}

\usepackage{graphicx}
\usepackage{dcolumn}
\usepackage{bm}

\usepackage[utf8]{inputenc}
\usepackage[T1]{fontenc}
\usepackage{etoolbox}

\newcommand*\G {\mathcal{G}}

\newcommand*\D {\mathcal{D}}
\newcommand*\F {\mathcal{F}}
\newcommand*\R {\mathbb{R} } 

\newcommand* \disp {\phi}
\newcommand* \atanG {\Phi}

\newcommand*\densityRatio {\Xi}
\newcommand*\distanceFunc {r}
\newcommand*\z {\varepsilon}

\newcommand*\flamethickness {l_f}
\newcommand*\flamespeed {s_f}
\newcommand*\channelWidth {\Lambda}
\newcommand*\cutoffWidth { \Lambda_{c} }
\newcommand*\kk {\kappa}

\newcommand* \RevisionDelete[1] {}

\makeatletter
\def\@email#1#2{%
 \endgroup
 \patchcmd{\titleblock@produce}
  {\frontmatter@RRAPformat}
  {\frontmatter@RRAPformat{\produce@RRAP{*#1\href{mailto:#2}{#2}}}\frontmatter@RRAPformat}
  {}{}
}%
\makeatother
\begin{document}

\preprint{}

\title{Parametric Learning of Time-Advancement Operators for Unstable Flame Evolution}
\author{Rixin Yu}
\email{rixin.yu@energy.lth.se}
\affiliation{ 
Department of Energy Sciences, Lund University, 22100 Lund, Sweden
}%

\author{Erdzan Hodzic}
\affiliation{%
Department of Manufacturing Processes, RISE Research Institutes of Sweden, 553 22 Jonkoping, Sweden 
}%


\date{\today}

\begin{abstract}
This study investigates the application of machine learning, specifically Fourier Neural Operator (FNO) and Convolutional Neural Network (CNN), to learn time-advancement operators for parametric partial differential equations (PDEs). Our focus is on extending existing operator learning methods to handle additional inputs representing PDE parameters. The goal is to create a unified learning approach that accurately predicts short-term solutions and provides robust long-term statistics under diverse parameter conditions, facilitating computational cost savings and accelerating development in engineering simulations. 
We develop and compare parametric learning methods based on FNO and CNN, evaluating their effectiveness in learning parametric-dependent solution time-advancement operators for one-dimensional PDEs and realistic flame front evolution data obtained from direct numerical simulations of the Navier-Stokes equations.
\end{abstract}

\maketitle


\section{Introduction \label{sec:intro}}

Resolving complex science and engineering problems often entails tackling nonlinear partial differential equations (PDEs), historically managed through numerical methods such as finite difference (FD), finite element (FE) or finite volume (FV) approaches. However, the computational demands of these methods have spurred the exploration of more efficient alternatives.

Recent advancements in machine learning and deep neural network methods, especially those adept at learning PDE solution operators, present promising solutions to these computational challenges. In the context of PDE solutions, operators play a crucial role, mapping one function to another. The focus of operator learning lies in methods capable of robustly learning PDE operators, providing excellent generalization performance.

Various operator learning methods have emerged in recent literature. An early notable method involves using deep convolutional neural networks (CNNs) \cite{CNN1,CNN2,CNN3,CNN4,CNN5,UNet,ConvPDE}, inspired by techniques from computer vision. These CNNs employ a finite-dimensional parameterization of the PDE operator, effectively mapping discretely represented functions between images. More recent advancements include a class of neural operator methods \cite{GraphKenerlNetwork, kovachki2021neural} capable of learning infinitely dimensional operators, exemplified by models like DeepOnet \cite{DeepONet} and Fourier Neuron Operator (FNO) \cite{FNO2020}. The theoretical underpinnings of these approaches have garnered support \cite{FNO_theroy, DeepONet_ChenChen_Theory, lanthaler2023nonlocal}, and both have demonstrated proficiency across a wide array of benchmark problems \cite{DeepONet_Nature, DeepONet_FNO_cmp}. Recent advancements have seen the further extension of neural operators, drawing inspiration from wavelet methods \cite{gupta2021multiwavelet, tripura2023wavelet}, and adapting methods for complex domains \cite{chen2023laplace}.

Of particular interest in the landscape of time-dependent PDE learning is the solution time-advancement operator. A proficiently learned operator in this regard can predict diverse instances of PDE solutions' evolution over extended durations, finding applications in tasks demanding detailed descriptions of turbulent flow dynamics. Examples range from weather forecasting\cite{FNOWeatherCast} to designing devices involving complex reacting flows.

While classical computational methods for turbulent problems are computationally intensive, the introduction of an operator learning method trained on the solution time-advancement operator holds the promise of enabling rapid predictions. This learned operator is expected not only to make accurate short-term predictions from varying initial conditions but also to project solutions over extended periods. However, the challenge of accurate long-term predictions in chaotic systems with nonlinear dynamics, sensitive to initial conditions, prompts an exploration of feasible alternatives. This often involves imposing weaker constraints, aiming for learned models that reproduce long-term statistics akin to ground truths.

In our recent study \cite{Yu2023}, three methods - CNN, FNO, and DeepOnet - were applied to learn the underlying operator governing the nonlinear development of unstable flame fronts in channels. Through recurrent training optimizing models for multiple consecutive predictions from a single input solution, it was found that FNO and CNN could effectively learn the front evolution. Specifically, FNO exhibited superior performance in capturing the intricate flame evolution in a wide channel, where the front evolves into cellular, fractal structures. On the other hand, CNN demonstrated better performance in predicting simpler flame evolution in a narrow channel, where the front evolves into a steady cusp shape. The channel-dependent front behavior is governed by PDEs, with the channel width as a parameter.

In the current work, the emphasis is on extending methods for learning time-advancement operators to include additional inputs of PDE parameters, capturing channel-dependent flame front evolution using a single neural network. The goal is to create a unified learning method capable of covering a spectrum of parameter values, providing accurate short-term solutions, and offering robust long-term statistics under varied parameter conditions. Given that engineering tasks often involve computational simulations over diverse parameter conditions, the development of a parametric operator learning method facilitates computational cost savings and accelerates development.

This work will concentrate on developing parametric learning methods based on two approaches - FNO and CNN. Integrating additional inputs into existing operator learning methods while retaining key mathematical structures presents a non-trivial challenge. Notably, the impact of added parameters on both short and long-term solutions must be incorporated to ensure better generalization of operator learning performance. It is worth noting that the influence of parameters on nonlinear systems may extend to bifurcation phenomena. For a detailed exploration of bifurcation behavior in machine learning studies, readers are referred to a paper \cite{ghadami2022deep} where a neural network archetype is constructed based on theories of the center manifold and normal forms.

The paper follows a structured organization: we commence with a description of the problem setup, followed by the presentation of parametric learning methods based on FNO and CNN. These methods will be compared in the context of learning parametric-dependent solution time-advance operators for two one-dimensional PDEs that model unstable front evolution due to distinct mechanisms of flame instability. Additionally, the methods will be showcased in their capacity to learn from realistic flame data obtained through direct numerical simulations of the Navier-Stokes equations. A summary and conclusion will be provided at the end.

\section{ Learning Problem setup  \label{sec:OLmethods} }

In this section, we describe the problem setup for learning a PDE operator, followed by the recurrent training methods.

Consider a system described by PDEs, which is often represented by multiple functions and maps between these functions. Here we consider a parametric operator mapping of
\begin{equation}
	\hat{\mathcal{G}}: 	\mathcal{V} \times \mathbb{R}^{d_\gamma}
 \to \mathcal{V}' 
\end{equation}
or
\begin{equation}
	\hat{\mathcal{G}}: 	(v(x) , \gamma ) \mapsto v'(x')
\end{equation}
which maps an input tuple of a function $v(x)$ and parameters $\gamma \in \mathbb{R}^{d_\gamma} $ into another function $v'(x')$. 
The input function is
\begin{equation}
	v: \D \to \R^{d_v}; x \mapsto v(x)
\end{equation}
where $ v\in \mathcal{V}$, and $\mathcal{V}= \mathcal{V}(\D;\R^{d_v})$
is a functional space with domain $\D \subset \R^d$ and codomain $\R^{d_v}$. 
The output function is 
\begin{equation}
v': \D' \to \R^{d_v'}; x'\mapsto v'(x')
\end{equation}
where $v' \in \mathcal{V}'$, and $\mathcal{V}'= \mathcal{V}'(\D';\R^{d_v'})$ is another functional space with domain $\D' \subset \R^{d'}$ and codomain $\R^{d_v'}$.

In this study, our main interest is in the solution time advancement operator with parametric dependence,  i.e.,
\begin{equation}
\hat{\mathcal{G}} : ( \disp(x,\bar{t}), \gamma ) \mapsto  \disp(x;\bar{t}+1) 
\label{eq:G_operator}
\end{equation}
where $\disp(x;\bar{t})$ is the solution to a PDE under certain parameters $\gamma$. Here $\bar{t}=t/\Delta_t \in \R$ is time normalized using a positive time increment  $\Delta_t$ (assumed to be small).
For simplicity, we consider autonomous problems and let $v'$ and $v$
share the same domain and codomain, i.e., $\D'=\D$, $\mathcal{V'}=\mathcal{V}$, $d'=d$, and $d_v'=d_v$. Furthermore, we assume simple (periodic) boundary conditions on $\D$ that do not vary with time.

Consider the approximation of the mapping $\hat{\mathcal{G}}$ using neural network methods. Let $\Theta$ be the space of all trainable parameters in the neural network. A neural network can be defined as a  map
\begin{equation}
 \mathcal{G}: \mathcal{V}\times \mathbb{R}^{d_\gamma} 
\times \Theta \to \mathcal{V}' \\ 
\text{ or equivalently }
 \mathcal{G}_{\theta}: \mathcal{V} \times \mathbb{R}^{d_\gamma}
 \to \mathcal{V}', \theta \in  \Theta. 
\end{equation}
Training neural network corresponds to finding a suitable choice of $\theta^*\in\Theta$ such that $\mathcal{G}_{ \theta^*} $  approximates $\hat{\mathcal{G}} $.

Starting from an initial solution function $\phi(x;t_0)$ and under fixed parameters values $\gamma$, the operator $\mathcal{G}_{\theta, \gamma}:= \mathcal{G}_\theta(\cdot,\gamma)$ can be recurrently applied by letting the input function being its output from previous prediction, this can roll out predicted solutions of arbitrary length. 
To learn a PDE solution advancement operator $\hat{\mathcal{G}}_\gamma:=\hat{\mathcal{G}}(\cdot, \gamma)$, 
 a well-trained neural network is expected to make accurate short-term predictions; while the long-term predicted solutions may not be precise when the PDE admits chaotic solutions, it is still preferable for the predicted solutions to share similar long-term statistics as ones in the PDE solutions.

Following our previous study \cite{Yu2023},  training of the recurrent network is set up in a one-to-many fashion, optimized for $\mathcal{G}_{\theta,\gamma}$ to make $n$ number of successive predictions from a single input function. 
More specifically, we denote by superscript $n$ the repeated application of an operator, either $\hat{\mathcal{G}}_\gamma$ or $\mathcal{G}_{\theta,\gamma}$,  e.g. $ \hat{\mathcal{G}}^n_\gamma :=  \underbrace{ \hat{\mathcal{G}}_\gamma \circ ... \circ \hat{\mathcal{G}}_\gamma }_{n}$.
Let all observation data be arranged in 1-to-$n$ pairs as 
$\left\{ v_{j}, (\hat{\mathcal{G}}_{\gamma_i}^1 v_j, \hat{\mathcal{G}}_{\gamma_i}^2 v_j,..., \hat{\mathcal{G}}_{\gamma_i}^n v_j ) \right\}_{j=1,i=1}^{Z_j,Z_i}$ 
where $v_j \sim \chi$ and $\gamma_i \sim \chi'$  are sequences drawn from two independent probability measures of $\chi$ and $\chi'$ respectively,
  the total number of input/output training pairs is  $Z=Z_i\times Z_j$.
Then, training the network $\mathcal{G}_\theta$ to approximate $\hat{\mathcal{G}}$ amounts to minimize 
\begin{equation} 
  \min_{\theta\in \Theta} \mathbb{E}_{ v \sim \chi, \gamma \sim \chi' }
 \left [ 
C (
  ( \mathcal{G}_{\theta,\gamma}^1 v,..., \mathcal{G}_{\theta,\gamma}^n v ) 
, 
(\hat{\mathcal{G}}_\gamma^1 v, ,..., \hat{\mathcal{G}}_\gamma^n v )
) 
\right ]
\label{eq:opt_1-to-n}
\end{equation}
where $C : \mathcal{V}^n  \times \mathcal{V}^n  \to \R $ is a cost function defined as relative mean square error(MSE), $\mathcal{V}^n$ denotes the Cartesian product of $n$ copies of $\mathcal{V}$.

It is noteworthy that the aforementioned one-to-many training setup is crucial for ensuring numerical stability in the learned solution advancement operator. As demonstrated in paper \cite{Yu2023}, deploying the same operator network but trained in a one-to-one setup frequently leads to divergent predictions upon repeated applications, attributed to unbounded error growth.

\section{Parametric operator learning Methods \label{sec:networks} }

In this section, we propose two models for learning the parametric operator of $\hat{\G}$.
 They are developed from two baseline methods, namely Convolutional Neural Network (CNN) and Fourier Neural Operator (FNO), designed to learn the non-parametric operator $\hat{\G}_\gamma$.

\subsection{Parametric convolutional neural network (pCNN)}

\begin{figure*}
	\centerline{
		\includegraphics[width=1\linewidth]{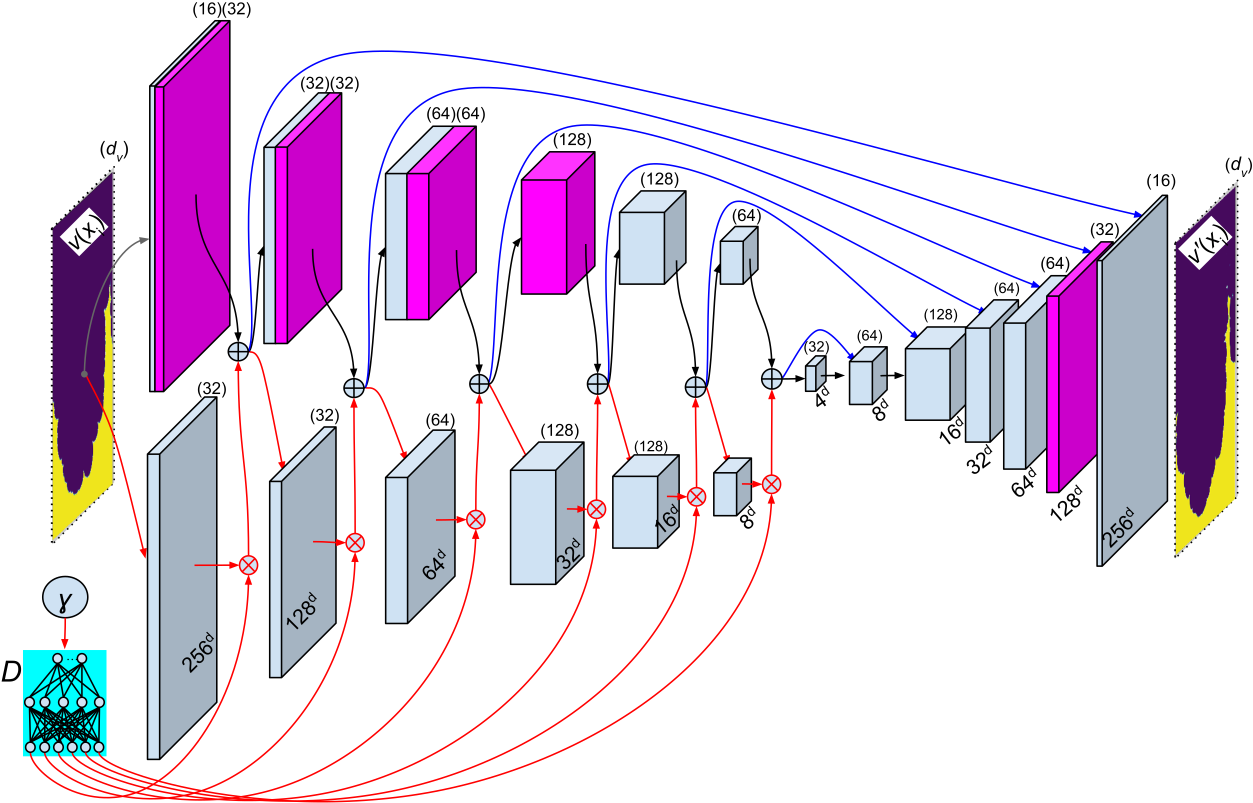}
	}
	\caption
	{
		\label{fig:CNN}
The parametric CNN is derived from the convolutional auto-encoder archetype by extending its encoder block, with the added components highlighted through linking with red lines. 
The sketch is demonstrated for a 2d input discretized as an image ($v(x_j)\in\R^{1\times 256\times 256}$) with $L$=6 levels of encoding. The channel number $c_l$ is shown on top in a bracket and the image size $N_l$ is on the bottom. Dashed lines refer to skip connection. Max-Pooling and upsampling are used to shrink and increase image size, respectively.  All gray blocks are implemented using a standard convolution layer (of filter size 3, stride 1, and periodic padding which enforces periodic boundary condition), and the magenta block is implemented by an Inception layer\cite{inception}. 
	}
\end{figure*}

The solution advancement operator $\hat{\G}_\gamma$, when discretized on an equispaced mesh, assumes the form of an image-to-image map. Image learning, akin to a computer vision task, is often accomplished through the application of deep CNN \cite{CNN1,CNN2,CNN3,CNN4,CNN5}.
The baseline architecture is a convolutional auto-encoder with skip connections, reminiscent of those employed in ConvPDE-UQ \cite{ConvPDE} and U-Net \cite{UNet}. This auto-encoder comprises an encoder block and a decoder block, with the input data undergoing successive transformations through a series of simple convolutional layers.

In this study, we propose a parametric CNN that expands the encoder block to account for parameter influence, as illustrated in Fig. \ref{fig:CNN} where the red lines highlight the new extensions.
Let $e^+_0$ denotes the input function $v(x_j)$ represented at a $x$-mesh, the encoder block is an iterative update procedure of
$(e^+_0,\gamma) \mapsto e^+_1 $, 
$(e^+_1,\gamma) \mapsto e^+_2 $,
$...$, $(e^+_{L-1},\gamma) \mapsto e^+_L$. 
At each level $l\in [0,...,L-1]$, the update $ (e^+_{l} , \gamma ) \mapsto e^+_{l+1}$ is achieved by a two-stage map  $ e^+_{l} \mapsto (e_{l+1},e^*_{l+1} ) \mapsto e^+_{l+1} $ with $ e^+_{l+1} = e_{l+1} + e^*_{l+1} \cdot D_{l}(\gamma)  $,  where the last function
$D_l: \mathbb{R}^{d_\gamma} \to \mathbb{R}$ converts the PDE parameters $\gamma$ into a scaling ratio. 
Here $e_l, e^*_l$ and $e^+_l \in \R^{c_l\times N_l} $  are three sequences of encoded `images', those images have $c_l$ channels and a total pixel number $N_{l} =  \prod_{k=1}^{d}  N_{l,(k)}  $ where $ N_{l,(k)} $ is pixel number in $k$-th dimension.

With an increased encoder level, images are recommended to increase in channel number (i.e. $c_{l} \leq c_{l+1}$, at least for small $l \leq 3$ ) but shrink in size ($N_{l,(k)}=2N_{l+1,(k)}$, except at first level where size do not change,i.e. $N_{0,(k)}=N_{1,(k)}$). 
The update $ e^+_{l} \mapsto (e_{l+1}, e^*_{l+1}) $  is decomposed into two sub-maps $ e^+_{l} \mapsto e_{l+1}$ and $e^+_{l} \mapsto e^*_{l+1} $, each of these two sub-maps is implemented by a size-2 max-pooling layer (to half the image size, but not needed at the first level $l=0$ ) followed by a few standard convolution layers (or replaced by Inception layer \cite{inception} for improved performance). Each of the above layers uses a filter size 3, periodic padding, and stride 1 followed by a ReLU activation.

Let $e'_L$ be the last encoded image $e^+_L$,
the decoder block is a reversed update procedure of 
$(e'_{L},e^+_{L-1}) \mapsto e'_{L-1}$,
$(e'_{L-1},e^+_{L-2}) \mapsto e'_{L-2}$, 
$...$,
$(e'_{2},e^+_{1}) \mapsto e'_{1}$.
Here $e'_l\in \R^{ c_l\times N_l}$ for $l\in[L-1,..,1]$ is a sequence of decoded images. 
Each decoder update  $(e'_{l+1},e^+_{l} )\mapsto e'_{l}$  is implemented by passing $e'_{l+1}$ through an up-sampling layer to double its image size then concatenating with $e^+_{l}$ along the channel dimension (i.e. the `skip' connections shown by the blue lines in fig. \ref{fig:CNN} ), followed by a few convolution layers. Nonlinear RELU activation is applied at each $l$ except at the last level  $l=1$ giving the final output $v'(x_j)=e'_1$.

\subsection{Parametric Fourier Neural Operator (pFNO) \label{sec:FNO} }

\begin{figure*}
	\centerline{
		\includegraphics[width=\linewidth]{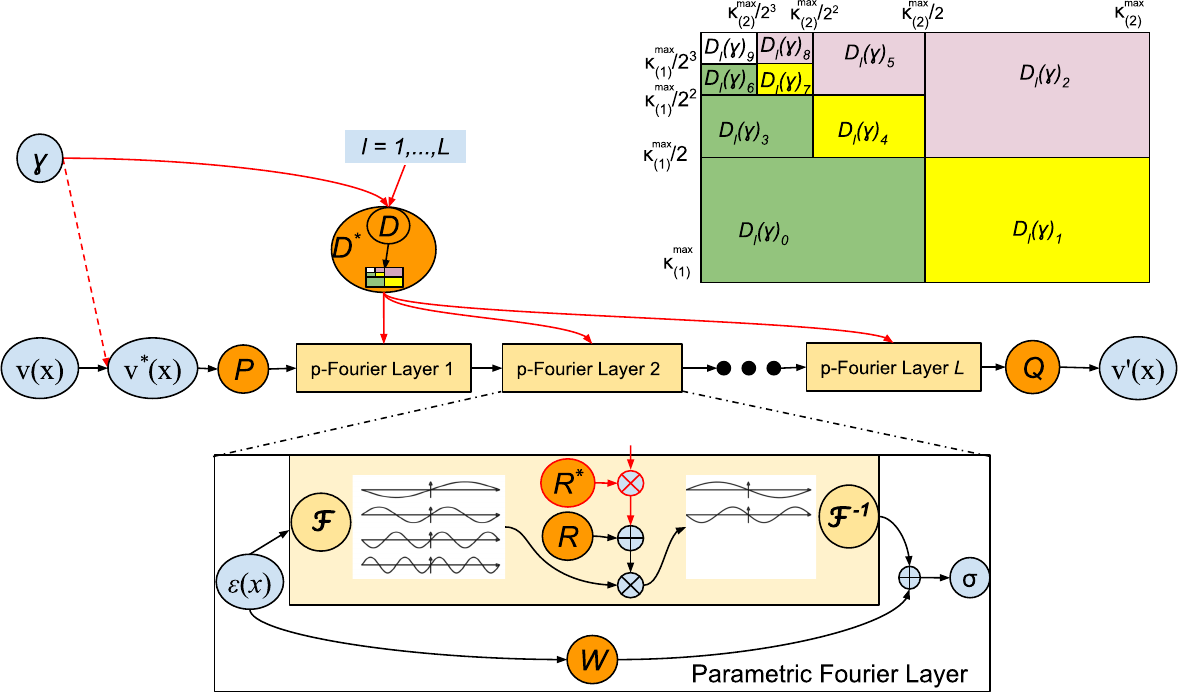}
	}
	\caption
	{
		\label{fig:FNO}
Illustration of the parametric extension of the Fourier Neural Operator (pFNO), highlighting the extended components with red solid lines connecting to the original FNO parts. The top-right inset provides a zoomed view of the second map inside the function $D^*$.
	}
\end{figure*}

In this section we describe the method for parametric extension of Fourier Neural Operator, starting from the baseline method. 

\subsubsection{Baseline FNO}

FNO\cite{FNO2020} is developed based on an early paper on neural operator  \cite{GraphKenerlNetwork} in which an infinite-dimensional operator is approximated by composing nonlinear activation and a class of integral kernel operators. 
In FNO the integral kernel is parameterized in Fourier Space.  
Similar to the pseudo-spectral method for solving nonlinear PDE, FNO involves intermediate data transformation alternatively switched in between Fourier space and physical space, as illustrated in fig. \ref{fig:FNO}. 

The main architecture of FNO is an iterative update procedure of 
$\z_1 \mapsto \z_2 \mapsto ... , \mapsto \z_L$, where $\z_l: \R^d \to \R^{d_\z}; x \mapsto \z_l(x)$ for $l=1,...,L$ is a sequence of functions. Here $\z_l(x)$ may be viewed as an `image' having $d_\z$ ($\geq1$) number of `color' channels if the $x$ domain is discretized on an equispaced mesh. 
Each update $\z_l \mapsto \z_{l+1}$ can be achieved by a Fourier Layer of
\begin{equation}
   \z_{l+1}  =    \sigma \left( \F^{-1} \{ \mathfrak{R}_l (  \F \{ \z_l \}  ) \}  +   W_l (\z_l)        \right)
	\label{eq:FNO}
\end{equation}
where $\sigma$ is a component-wise nonlinear activation function(e.g. ReLU) and the function $W_l: \R^{d_\z} \to \R^{d_\z}; \z_l \mapsto W_l( z_l )$ performs a channel-wise linear transformation which can be parameterized by a single convolutional neural network layer with kernel size 1. 

The remained operation in  Eq. \eqref{eq:FNO} starts with Fast Fourier Transform(FFT) $\F$ and ends with its inverse  $\F^{-1}$. 
First, on the complex-valued Fourier modes $\F \{\z_l\} $ we truncate higher frequency modes than $\kappa^{max}= \Pi_{i=1}^{d} \kappa_{(i)}^{max}$
to obtain  $\F \{\z_l\} \in  \mathbb{C}^{\kappa^{max} \times d_\z}$, here $\kappa_{(i)}^{max}$ denotes the number of modes kept in the $i$-th dimension.
Then, these truncated modes are linearly transformed by a function 
$\mathfrak{R}_l: \mathbb{C}^{\kappa^{max} \times  d_\z} \to \mathbb{C}^{\kappa^{max} \times d_\z } $ as,
\begin{eqnarray}
 \mathfrak{R}_l (\F \{\z\} ) _{\kappa,i} = 
 \sum_{j=1}^{d_\z}    
   (R_l)_{\kappa,i,j} 
\F \{\z\}_{\kappa,j}  ,  \nonumber \\  
\kappa=1,...,\kappa^{max} \text{ and }  i=1,...,d_\z
 \label{eq:FNO_R}
\end{eqnarray}
where $R_l \in  \mathbb{C}^{\kappa^{max} \times d_\z \times d_\z}$ is a trainable weight tensor. 
Lastly, the inverse Fourier Transform $\F^{-1}$ brings the modified Fourier modes back to physical space.

For the input function $v(x)\in \R^{d_v}$ to be fed into above Fourier layers,  a local transformation  
$P: \R^{d_v} \to \R^{d_\z}$ is required to lift the input  to a higher dimension (i.e. $\z_0(x) = P( v(x) ) $  and $d_\z\geq d_v$ ),  
Similarly, $\z_{L} \in \R^{d_\z}$ outputted from the last Fourier layers needs to be projected to become a lower dimension output function of $ v'(x) = Q( \z_L(x) )$,  
here $Q: \R^{d_\z} \to \R^{d_v} $ is another local transformation. Both $P $ and $Q$ can be parameterized using simple networks such as multiple layer perceptions(MLP).

FNO has several pleasant features: (i) FNO is mesh-invariant and can be trained at low resolution and then evaluated at high resolutions. 
(i) FNO can handle different boundary conditions through $W_l$, while particularly with periodic conditions FNO is invariant to a translation in the input function, i.e.  
 i.e., $\G_{\gamma; \theta} ( v(x+a) )(x) = \G_{\gamma; \theta} ( v(x) )(x+a) $ for any $a \in \D$.
(iii)  FNO can also be promoted to learn the higher dimensional map  $\hat{\mathcal{G}}^*$ where both input and output functions are extended to have a time dimension.
One restriction of FNO attributes to the usage of FFT,
therefore both the input and output functions $v$ and $v'$ should be discretized on an equispaced mesh. Theoretical analysis on error bounds of FNO method is in paper \cite{FNO_theroy}.

\subsubsection{Parametric extension \label{sec:pFNOExtension} }

To incorporate parametric influence into FNO, we enhance the baseline Fourier layer by introducing additional learnable complex weights that encompass parametric influences across different ranges of wave numbers. The primary modificationa are illustrated by the components connected by red solid lines in Fig. \ref{fig:FNO}.

Letting $v^*(x)=v(x)$, the Fourier layer is modified as a parametric-dependent update procedure of:
$(\z_1,\gamma) \mapsto \z_1$ ,
$(\z_2, \gamma) \mapsto \z_2$,
$...$,
$(\z_{L-1}, \gamma)  \mapsto \z_L$. 
Each update $(\z_l, \gamma) \mapsto \z_{l+1}$ can be achieved by Eq. \eqref{eq:FNO} 
through replacing $\mathfrak{R}_l:\mathbb{C}^{\kappa^{max} \times  d_\z}  \to \mathbb{C}^{\kappa^{max} \times  d_\z}$ by a different function $\mathfrak{R}^*_l: \mathbb{C}^{\kappa^{max} \times  d_\z}  \times \R^{d_\gamma} \to \mathbb{C}^{\kappa^{max} \times  d_\z} $. 
A parametric Fourier layer becomes
\begin{equation}
   \z_{l+1}  =    \sigma \left( \F^{-1} \{ \mathfrak{R}^*_l (  \F \{ \z_l \} , \gamma ) \}  +   W_l (\z_l)        \right)
	\label{eq:pFNO}
\end{equation}
with
\begin{eqnarray}
 \mathfrak{R}^*_l (\F\{\z\} , \gamma )_{\kappa,i} =   
     \sum_{j=1}^{d_\z} [ (R_l)_{\kappa,i,j}   +(R^*_l)_{\kappa,i,j}   D_l^*(\gamma)_{\kappa}  ] \F \{\z\}_{\kappa,j} , \nonumber \\
\kappa=1,...,\kappa^{max} \text{ and }  i=1,...,d_\z
\label{eq:Rstar}
\end{eqnarray}
where
$R^*_l \in \mathbb{C}^{\kappa^{max} \times  d_\z \times d_\z} $ is an additional  trainable complex weight tensor
  and
 $D^*_l: \R^{d_\gamma} \to \R^{\kappa^{max}} $ is a function which maps the parameters $\gamma$ to $\kappa^{max}$-number of positive ratios (for re-scaling $R^*_l$). 
 $D^*_l$ can be realized by composing two functions: a first function $D_l: \R^{d_\gamma} \to \R^{N_D}$ outputs $N_D= 1+(2^{d}-1)\cdot N_\gamma $ number of values and it is parameterized by a shallow MLP; the second function performs the mapping of $\R^{N_D}\to  \R^{\kappa^{max}} $. 
Following a similar approach to constructing pCNN, this second mapping function is implemented hierarchically, enabling the distribution of parameter influence across different ranges of wave numbers.
More specifically, at one dimension this map becomes  
$   D^*_l(\gamma) _\kappa =    D_l(\gamma) _i$
 for $\kappa \in  ( \frac{ \kappa^{max}}{2^{i+1}} , \frac{ \kappa^{max}}{2^{i}} ] $  at $i =0,..,N_\gamma-1$ , 
 and, for $ \kappa  \in  (0, \frac{ \kappa^{max}}{2^{N_\gamma}}] $  at $ i =N_\gamma$.
At two dimensions the map reads as
   $(D^*_l(\gamma) )_{\kappa_{ (1) }, \kappa_{ (2)} }=    (D_l(\gamma) )_{3i+j} $,  
for any $i =0,..,N_\gamma-1$ together with one of the following three conditions   
 (i) $j=0$ and $\kappa_{(1)},\kappa_{(2)} = (0, \frac{ \kappa_{(1)}^{max}}{2^{i+1}}] \times  (\frac{ \kappa_{(2)}^{max}}{2^{i+1}} , \frac{ \kappa_{(2)}^{max}}{2^{i}} ]  $,  
 (ii)  $j=1$ and $ \kappa_{(1)},\kappa_{(2)} = ( \frac{ \kappa_{(1)}^{max}}{2^{i+1}} , \frac{ \kappa_{(1)}^{max}}{2^{i}} ] \times  (\frac{ \kappa_{(2)}^{max}}{2^{i+1}} , \frac{ \kappa_{(2)}^{max}}{2^{i}} ]    $, 
  (iii)  $j=2$  and   $\kappa_{(1)},\kappa_{(2)} = ( \frac{ \kappa_{(1)}^{max}}{2^{i+1}} , \frac{ \kappa_{(1)}^{max}}{2^{i}} ] \times  [ 0 , \frac{ \kappa_{(2)}^{max}}{2^{i+1}} ] $. 
%
A 2D example with the hyperparameter $N_\gamma=3$ is shown in the upper right corner in fig. \ref{fig:FNO}.

\subsubsection{Variants (pFNO*) \label{sec:variant}}
An alternative approach to incorporate parameter influence into the baseline FNO method involves appending each parameter value $\gamma \in \mathbb{R}^{d_\gamma}$ to the codomain of the input function $v(x)$. This modification results in a different input function $v^*: \mathbb{R}^d \to \mathbb{R}^{d_v + d_\gamma}$ (in contrast to letting $v^*(x) = v(x)$ in the previous section), represented by the red dashed line in Figure \ref{fig:FNO}. 
This modification is a simple extension to the baseline FNO and will be referred to as pFNO* for future reference.

\section{ Results and discussions \label{sec:numericalexp} }

The two models,  namely pFNO and pCNN, as detailed in Section \ref{sec:networks}, are deployed to learn the parametric-dependent operator map governing the evolution of unstable flame fronts. Two distinct datasets are utilized to train these models. The first dataset is derived from 1D solutions of two modeling equations: the 1D Michelson-Sivashinsky (MS) equation \cite{michelson1977nonlinear, SivaEq} and the 1D Kuramoto-Sivashinsky (KS) equation \cite{kuramoto1978diffusion, SivaEq}. These equations capture intrinsic flame instability arising from Darrieus-Landau \cite{DARRIEUS1938UNPB,landau1988theory} and diffusive-thermal \cite{zeldovich1944selected, sivashinsky1977diffusional} mechanisms, respectively. Although the front solutions from these equations are confined to 1D functions (dependent along the channel wall-normal direction), a substantial number of solution sequences can be generated, varying initial conditions and parametric values.

The second dataset encompasses 2D front solutions \cite{YBB15PRE} obtained from direct numerical simulations (DNS) of reacting Navier-Stokes equations. The DNS provides more realistic and intricate front solutions that cannot be represented by 1D functions. However, due to computational demands, the 2D dataset is limited in the number of solutions available for training.

In the subsequent sections, we evaluate the performance of our proposed models for parametric operator learning. Initially, we compare their effectiveness using the more extensive 1D dataset. Subsequently, we showcase the utilization of the 2D-version models for training on the DNS dataset.

\subsection{1D governing equations and training dataset \label{sec:1d}}

Consider modeling unstable flame front development in a periodic channel. Let $x$ represent a normalized spatial coordinate along the channel's wall-normal direction, i.e., $x \in \D = [-\pi, \pi)$, and $t$ denotes time. Introduce a displacement function $\disp(x,t): \R \times \R \to \R$ describing the stream-wise coordinate of a zero-thickness flame front undergoing Darrieus-Landau instability. This evolution can be captured by the Michelson-Sivashinsky (MS) equation \cite{michelson1977nonlinear, SivaEq}:
\begin{eqnarray}
\partial_t \disp + \frac{1}{2} (\partial_x \disp)^2 = \nu \partial_{xx}^2 \disp + \Gamma(\disp) \label{eq:MS}
\end{eqnarray}
Similarly, the flame front evolution due to diffusive-thermal(DT) instability is modeled by the Kuramoto-Sivashinsky (KS) equation \cite{kuramoto1978diffusion, SivaEq}:

\begin{eqnarray}
\partial_t \disp + \frac{1}{2 \beta^2} (\partial_x \disp)^2 = -\frac{1}{\beta^2} \partial_{xx}^2 \disp - \frac{1}{\beta^4} \partial_{xxxx}^4 \disp \label{eq:KS}
\end{eqnarray}
Both equations have periodic boundary conditions and initial conditions of $\disp(x,0)=\disp_0\in L^2_{\text{per}}([-\pi,\pi))$. In Eq. \eqref{eq:MS}, $\Gamma: \disp(x) \mapsto -\mathcal{H}(\partial_x \disp)$ is a linear singular non-local operator defined using the Hilbert transform $\mathcal{H}$, or in terms of spatial Fourier transform, denoted by $\F_k(\disp(x))$ and its inverse $\F^{-1}$:

\begin{equation}
\Gamma: \disp(x) \mapsto \F^{-1} (|\kk| \F_\kk (\disp(x))). \label{eq:Lmap}
\end{equation}

Equations \eqref{eq:MS} and \eqref{eq:KS} include positive parameters $\nu$ and $\beta$, where $\nu$ depends on channel width and gas thermal expansion, while $\beta$ depends on the disparity between reactant mass and heat diffusion. The KS equation is known to exhibit chaotic solutions at large $\beta$ and is often used as a benchmark case for PDE learning studies. The MS equation, less known outside the flame instability community, can be exactly solved using a pole-decomposition technique \cite{Thual_Frisch_Henon_poledecomp}, transforming it into a set of ODEs with finite freedoms. Additionally, at large $\nu$, the MS equation admits a steady solution in the form of a giant cusp front. However, at smaller $\nu$, the equation becomes sensitive to noise, resulting in unrest solutions with everlasting small wrinkles atop a giant cusp. Further details about known theory can be found in \cite{Vaynblat_matalon_polestability1,
Vaynblat_matalon_polestability2,Olami_noise,denet2006stationary,Kupervasser_pole_book,
Karlin2002cellular,Creta2020propagation,CRETA2011INST,Yu2023}.

The 1D equations \eqref{eq:MS} and \eqref{eq:KS} are solved using a pseudo-spectral approach along with a Runge-Kutta (4,5) time integration method. All numerical solutions are obtained on a uniformly spaced 1D mesh of 256 points. For the MS Equation \eqref{eq:MS}, training solutions are generated at six parameter values of $\nu \in [0.025, 0.035, 0.05, 0.07, 0.1, 0.15]$. Given that, at large values of $\nu$, the long-term MS solution tends to evolve into a nearly stationary single cusp front, 250 sequences of consecutive solutions are generated for each of the three large $\nu \in [0.07, 0.1, 0.15]$. Each sequence contains 1000 consecutive solutions separated by a time interval of $\Delta_t=0.015$ and starts from random initial conditions $\disp_0(x)$ sampled from a uniform distribution over [0, 0.03]. For smaller $\nu \in [0.025, 0.035, 0.05]$, the training data is similarly created, but adjustments are made to better represent the noise-affected unrest solutions. Specifically, 250 random sequences of solutions are generated over a shorter duration $0 < t < 7.5$, each containing 500 consecutive solutions at intervals of $\Delta_t$. The long-time solution behavior is represented by one extra-long sequence, including 125,000 consecutive solutions throughout $0 < t < 1875$.

For the KS Equation \eqref{eq:KS}, the training dataset comprises 250 randomly initialized solution sequences over $0 < t < 7.5$ for each of five parameter values of $\beta \in [6, 9, 12, 18, 24]$. The KS solutions at all these $\beta$ values remain chaotic. For both MS and KS equations, the validation dataset is similarly created but contains only 10 percent of the data in the training dataset.

\subsection{ Learning 1D flame evolution}

\begin{table}
	\caption{  Relative $L^2$ train/validation errors for all 1D operator networks \label{Table1} }  
	\centerline{
		\begin{tabular}{ccccccccc}  
			\hline
			   Model & Governing Eq. & Parameter range &	Train $L^2$	& Valid. $L^2$	\\
			\hline
			pCNN   & MS& $\nu\in [0.025 - 0.15]$    &  0.0167   &  0.0177 \\ 
			pFNO*  & MS& $\nu\in [0.025 - 0.15]$   &    0.0180  & 0.0186    \\ 
			pFNO & MS& $\nu\in [0.025 - 0.15]$  &  \textbf{0.0101}    &  \textbf{0.0105} \\ 
			\hline
			pCNN   & KS & $\beta\in [6-24]$    &  0.0189   &  0.0200 \\ 
			pFNO*  & KS & $\beta\in [6-24]$   &    0.0173  & 0.0181    \\ 
			pFNO  & KS & $\beta\in [6 -24]$  &  \textbf{0.0102}    &  \textbf{0.0108}\\ 
  			\hline
		\end{tabular}
	}
\end{table}

The training datasets described in the previous section are employed to learn two parametric solution advancement operators, denoted as $\hat{\mathcal{G}} : (\phi(x;t), \gamma) \mapsto \phi(x;t+\Delta_t)$. These operators correspond to the MS equation \eqref{eq:MS} with $\gamma =\nu$ and the KS equation \eqref{eq:KS} with $\gamma=\beta$, representing unstable front evolution due to the DL and DT instability mechanisms, respectively. Subsequently,  these operators are referred to as $\hat{\mathcal{G}}_\nu^{DL}$ and $\hat{\mathcal{G}}_\beta^{DT}$.

In the current study, two parametric learning models, pFNO and pCNN, introduced in section \ref{sec:OLmethods} are applied to learn  $\hat{\mathcal{G}}_\nu^{DL}$ and $\hat{\mathcal{G}}_\beta^{DT}$.  
For comparison, we also include a third model pFNO* (a simple variant of FNO through appending the parameter $\gamma$ to the codomain of input function, described in section \ref{sec:variant} ).
Given a fixed parameter value, the learned operator is expected to make recurrent predictions of solutions over an extended period. 
The training for such operators aims not only for accurate short-term predictions but also for robust predictions of long-term solutions with statistics similar to the ground truth.
 
As demonstrated in a previous study \cite{Yu2023}, achieving this involves organizing the training data in a 1-to-n pair (n=20 for 1D dataset), as expressed in Eq. \eqref{eq:opt_1-to-n}, optimized for accurately predicting 20 successive steps of outputs from a single input over a range of parameter values.

\begin{figure*}
	\centerline{
	\includegraphics[width=1\linewidth]{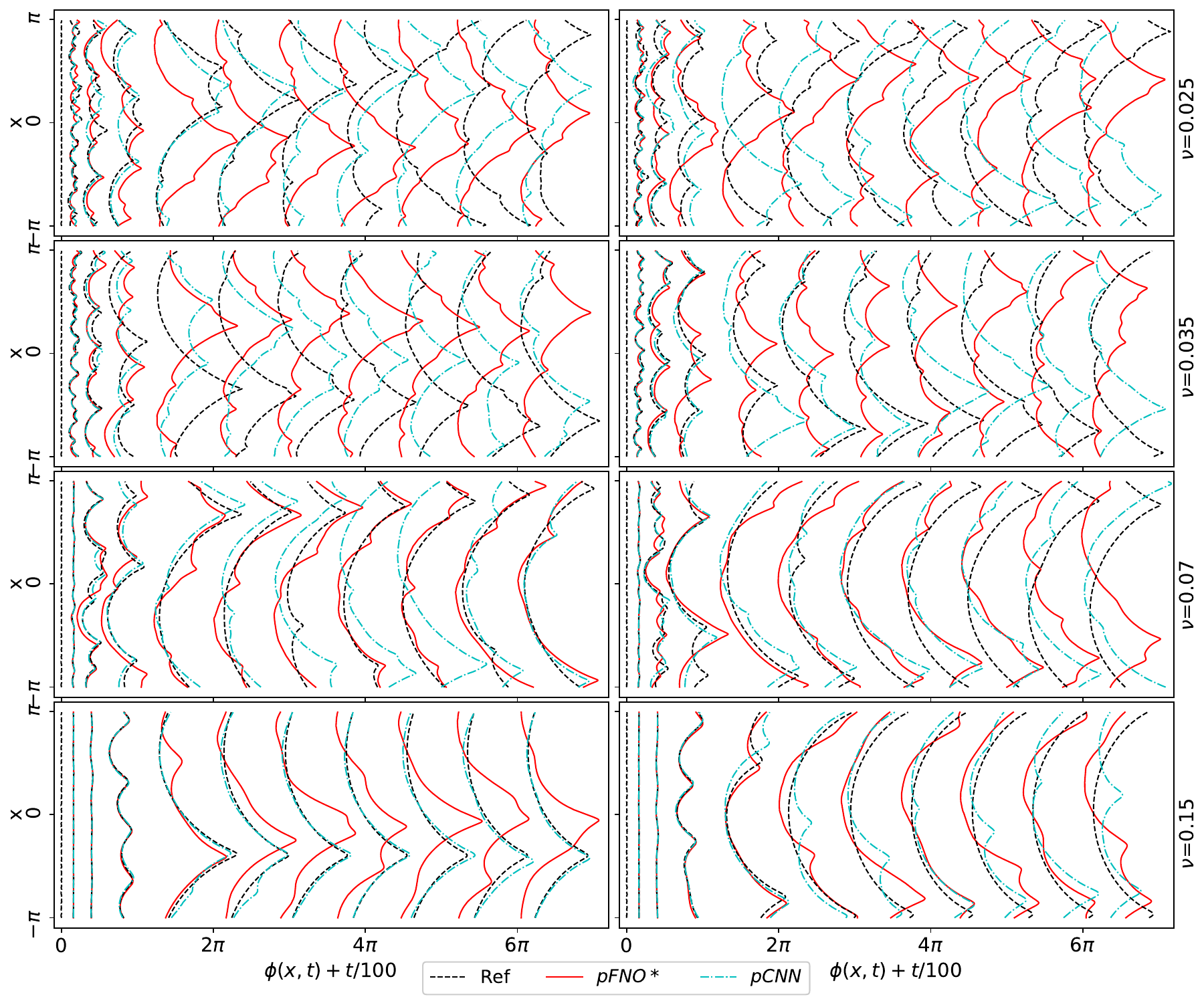}
	}
	\caption
	{
		\label{fig:disp_MS}
Long-term solutions of the 1D MS equation \eqref{eq:MS} for front displacement $\disp(x,t)$ at four different parameters $\nu \in [0.025,0.035,0.07,0.15]$ (from top to bottom row). Reference solutions obtained using high-order numerical methods are represented by black dashed lines, while predictions from parametric operator learning methods pFNO* and pCNN are denoted by red solid and cyan long-dash lines, respectively. Each pair of solution sequences, displayed in the left and right columns, starts from differently randomized initial fronts. The sequences include eleven snapshots of $\disp(x,t_j)$ with $t_j=j\Delta_t$ at $j\in[0,50,125,250,500,750,1000,1250,1500,1750,2000]$ and a fixed time interval $\Delta_t$=0.015. A time shift ($t/100$) is applied to displayed fronts to reduce overlap.
	}
\end{figure*}

\begin{figure*}
	\centerline{
		\includegraphics[width=1\linewidth]{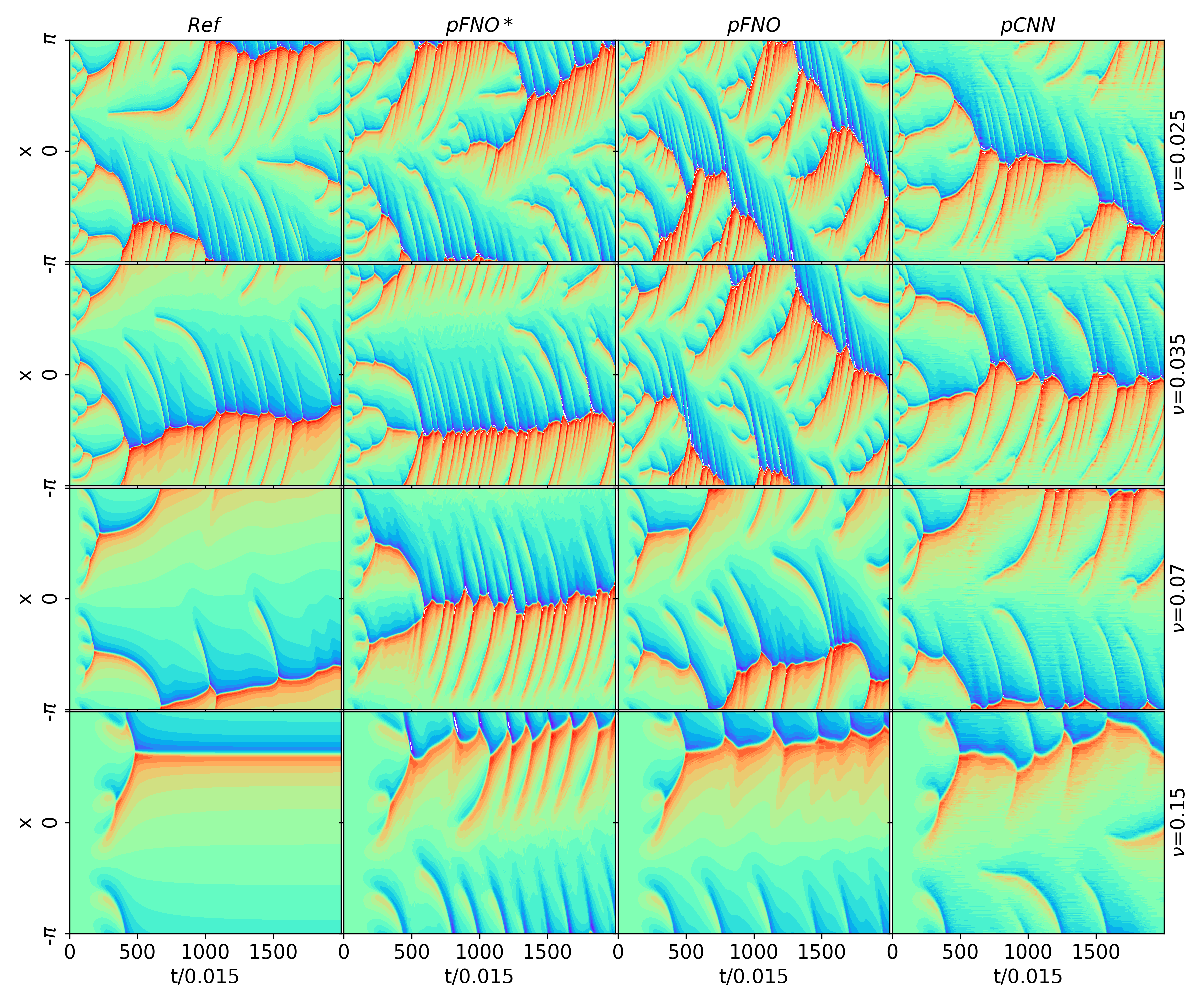}
   }
	\caption{
		\label{fig:uSlope_MS}
Front slope of $\partial_x \disp(x,t)$ calculated from a single instance reference solution (1st column) of the 1D MS equation \eqref{eq:MS} (at four different parameters $\nu \in [0.025,0.035,0.07,0.15]$, from top to bottom row) is compared against the predictions by pFNO*, pFNO, and pCNN shown in the 2nd, 3rd, and 4th column, respectively. The rainbow color map represents values from negative to positive.
	}
\end{figure*}

\begin{figure*}
	\centerline{
		\includegraphics[width=1\linewidth]{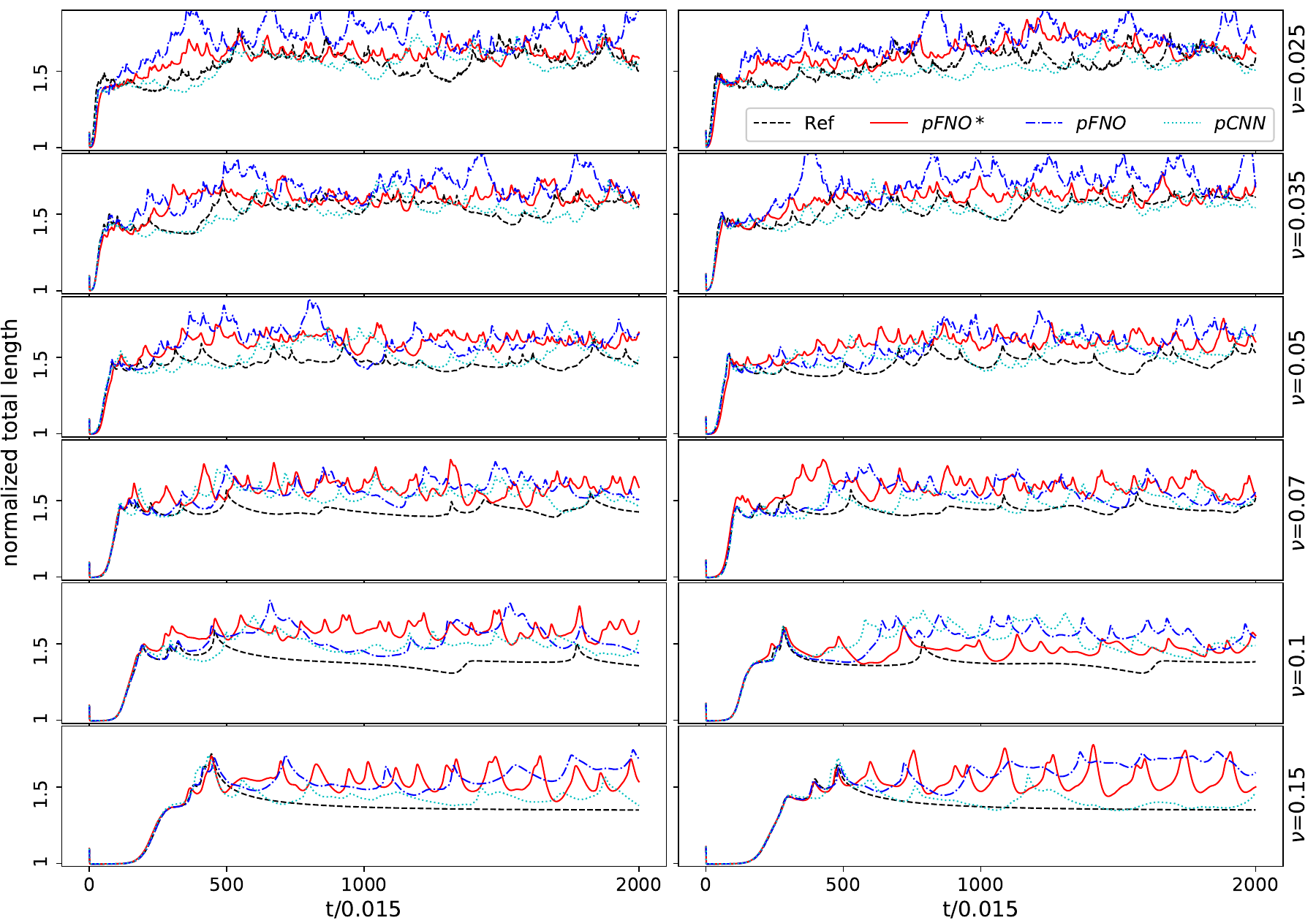}
   }
	\caption{
		\label{fig:len_MS}
Normalized total front length (i.e. $\int_\D (\disp_x^2 +1 )^{1/2} dx /\int_\D dx $) calculated from two random instances(shown in left and right column respectively) of reference solutions(black dash line) to the 1d MS equation (for each of six parameter values $\nu$ shown from top to bottom row) are compared against the corresponding predictions by three networks of pFNO* (red solid line), pFNO(blue dash-dot line) and pCNN (cyan dash line). 
	}
\end{figure*}

\begin{figure*}
	\centerline{
		\includegraphics[width=.7\linewidth]{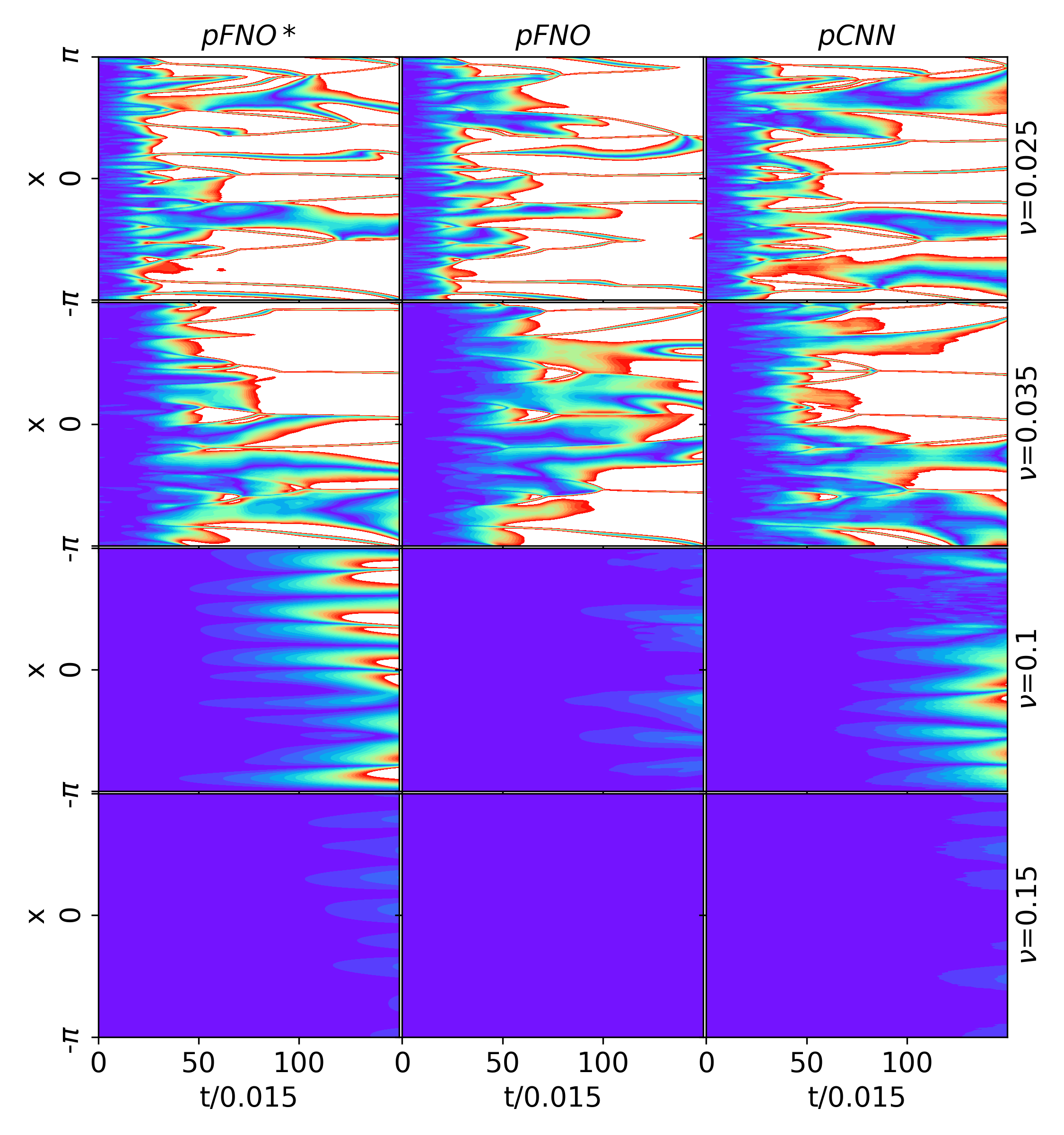}
   }
	\caption{
		\label{fig:err_MS}
Time evolution of the relative $L^2$-error between the reference solution of the 1D MS equation, initialized with a random condition at each of four parameter values of $\nu$, and the corresponding predictions by three networks: pFNO*, pFNO, and pCNN. Rainbow colors from blue to red represent values ranging from 0 to 0.1; values above 0.1 are truncated and displayed as white.
	}
\end{figure*}

\begin{figure*}
	\centerline{
		\includegraphics[width=1\linewidth]{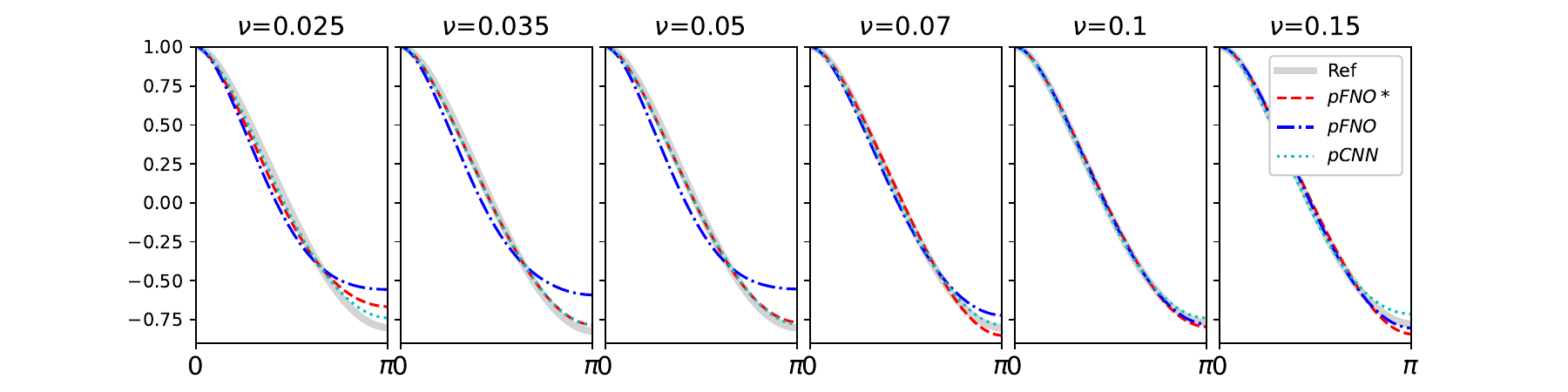}
   }
	\caption{
		\label{fig:corr_MS}
	Comparison of the auto-correlation function $\mathcal{R}(r)$, measuring the long term reference solutions to 1d MS equation at different parameter values $\nu$, against the corresponding predictions using three networks: pFNO*, pFNO and pFNO.
	}
\end{figure*}

\begin{figure*}
	\centerline{
	\includegraphics[width=1\linewidth]{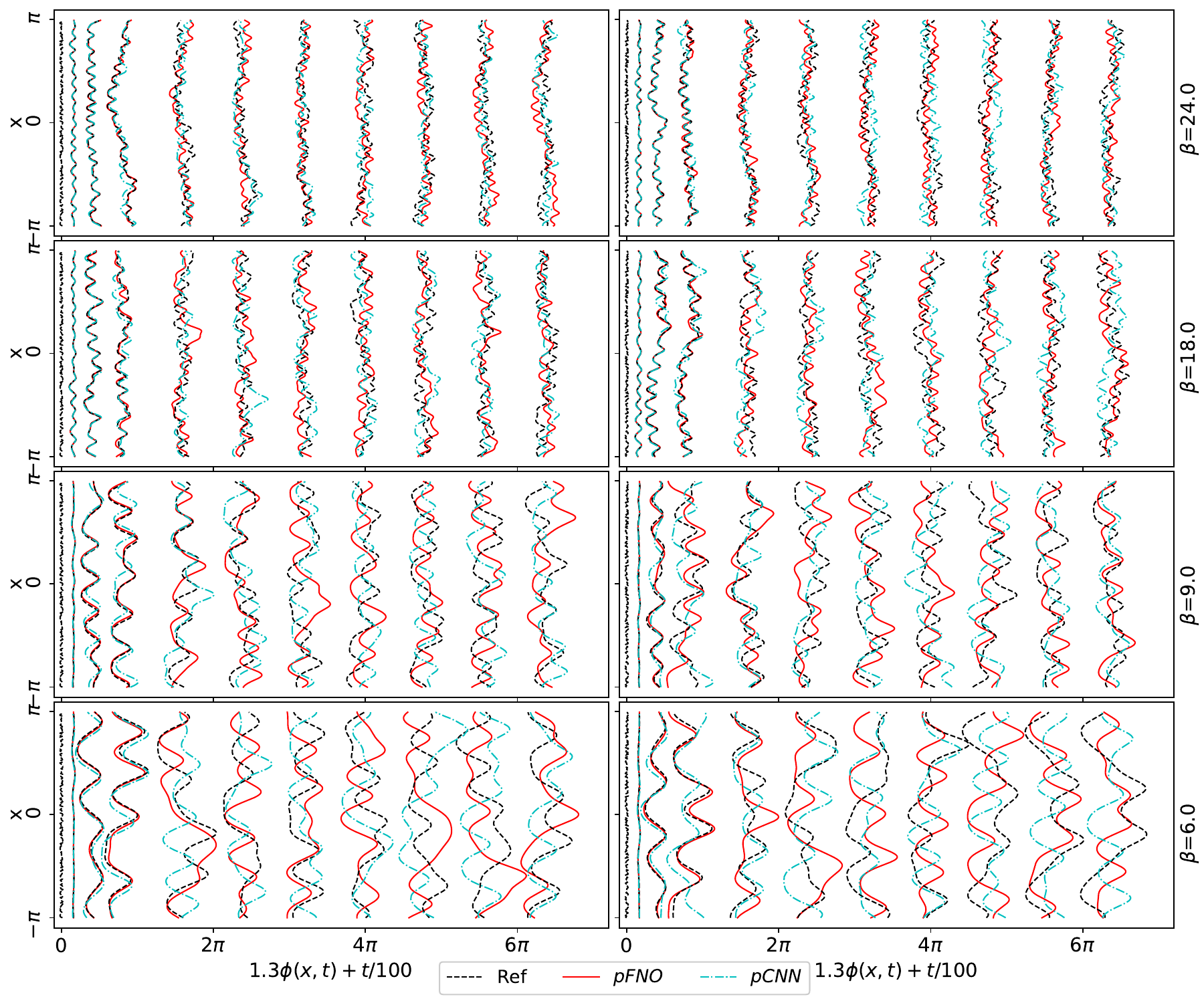}
	}
	\caption
	{
		\label{fig:disp_KS}
Long-term solutions of the 1D KS equation \eqref{eq:KS} at four different parameters $\beta \in [24,18,9,6]$ (from top to bottom row), obtained by high-order numerical methods as a reference, are compared against predictions by pFNO and pCNN. Other details remain consistent with those in Figure \ref{fig:disp_MS}.
	}
\end{figure*}

\begin{figure*}
	\centerline{
		\includegraphics[width=1\linewidth]{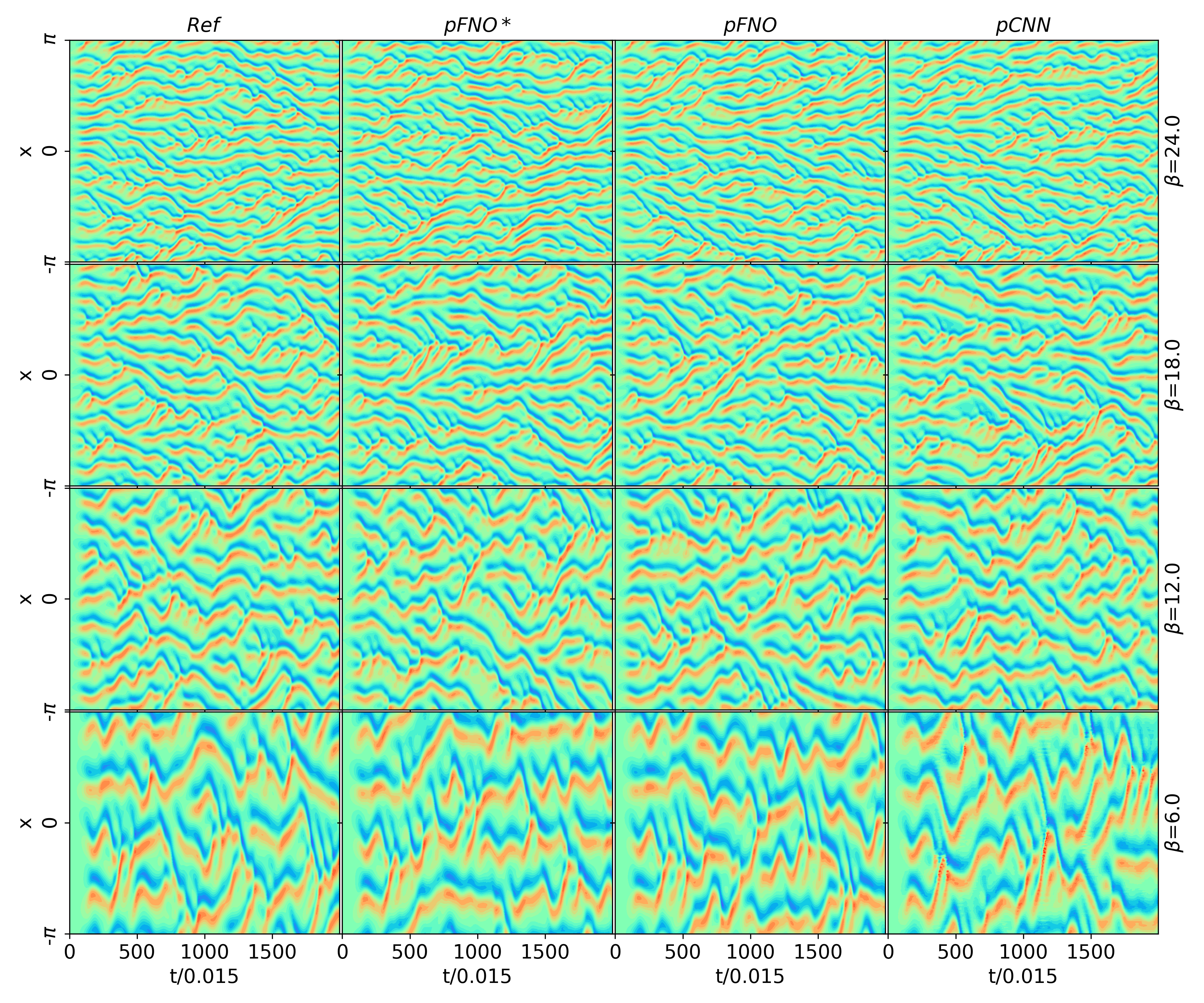}
   }
	\caption{
		\label{fig:uSlope_KS}
Comparison of front slopes between reference solutions to the 1D KS equation at four parameters $\beta$ = 24,18,12,6 and predictions by pFNO*, pFNO, and pCNN. The format and details remain consistent with those presented in Figure \ref{fig:uSlope_MS}.
	}
\end{figure*}

\begin{figure*}
	\centerline{
		\includegraphics[width=1\linewidth]{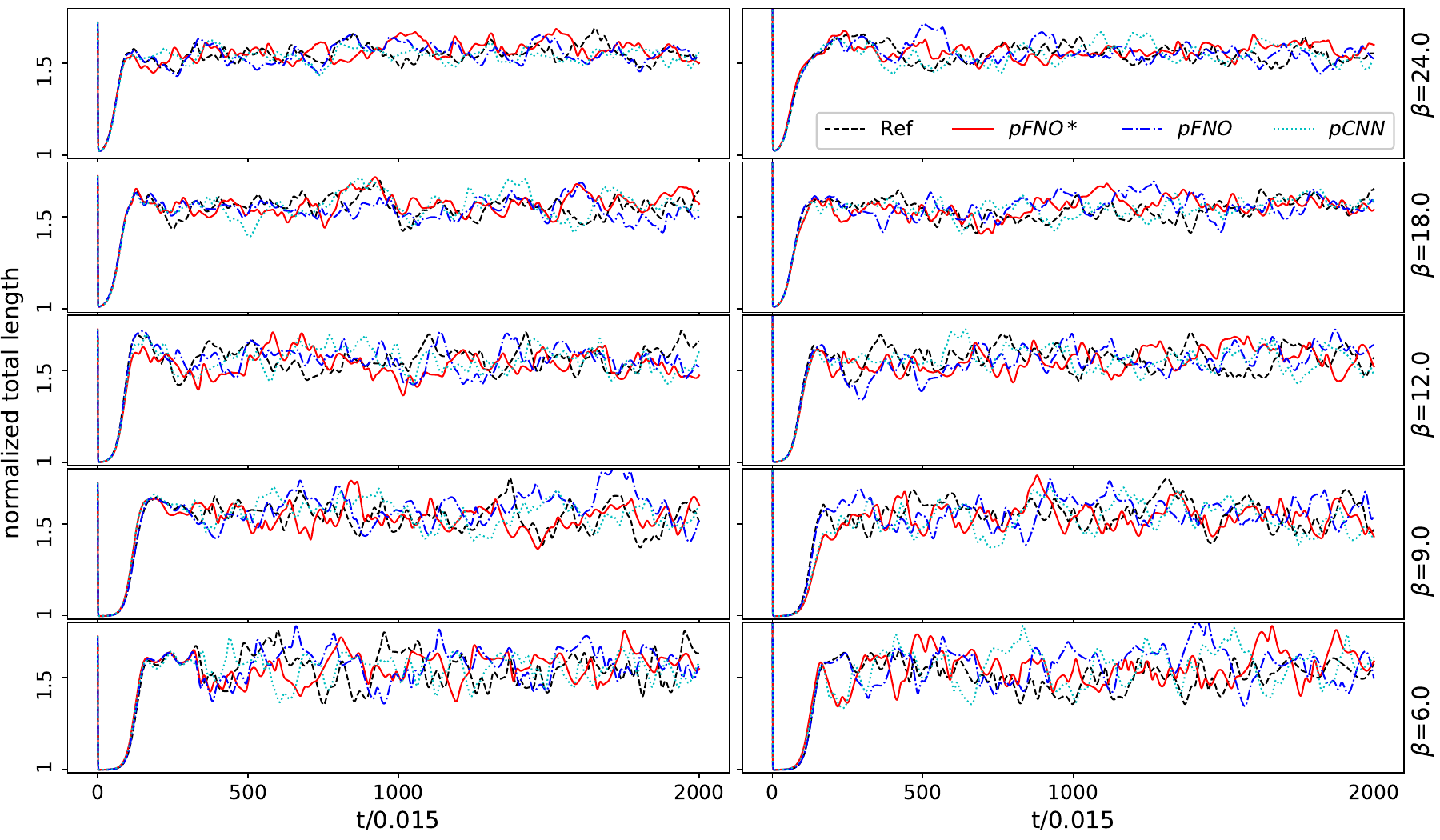}
   }
	\caption{
		\label{fig:len_KS}
Comparison of normalized total front length for 1d KS equation at  different parameters $\beta$ obtained from reference solution and predictions by pFNO*, pFNO and pCNN. Other details can be found in fig. \ref{fig:len_MS}.
	}
\end{figure*}

\begin{figure*}
	\centerline{
		\includegraphics[width=0.7\linewidth]{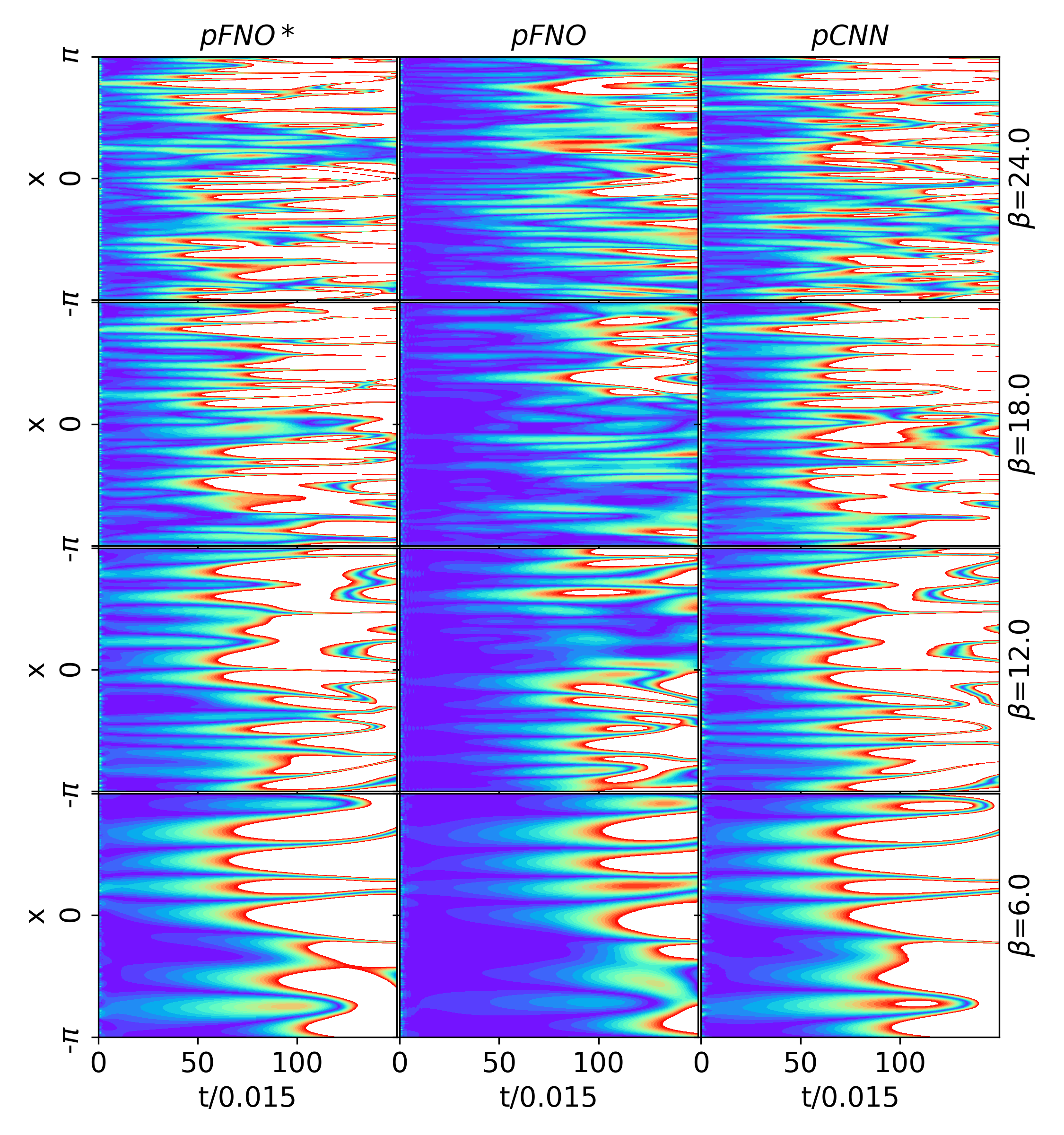}
   }
	\caption{
		\label{fig:err_KS}
	Relative $L2$-error between reference solution of 1d KS equation and the corresponding predictions by three networks of pFNO*, pFNO and pCNN. Other details are the same as in fig. \ref{fig:err_MS}.
	}
\end{figure*}

\begin{figure*}
	\centerline{
		\includegraphics[width=1\linewidth]{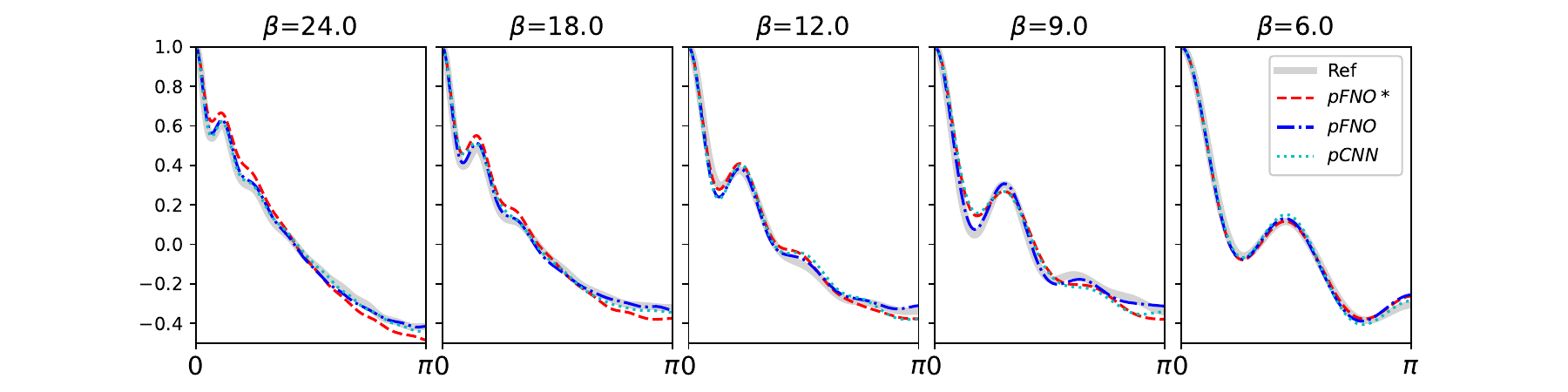}
   }
	\caption{
		\label{fig:corr_KS}
	Comparison of the auto-correlation function $\mathcal{R}(r)$, measuring the long-term reference solutions to 1d KS equation at different parameter values $\beta$, against the corresponding predictions using three networks: pFNO*, pFNO and pFNO.
	}
\end{figure*}

Table \ref{Table1} presents the training/validation (relative $L^2$) errors for the obtained models on the two equations. Additional details on training and model hyper-parameters are provided in Appendix \ref{app:nn_detail}.
For learning $\hat{\mathcal{G}}_\nu^{DL}$,
 Fig. \ref{fig:disp_MS} compares two extended sequences of front displacements predicted by all models against the reference ones. Figs. \ref{fig:uSlope_MS} and \ref{fig:len_MS} illustrate analogous comparisons for the front slope of $\disp_x$ and normalized total front length of $\int_\D (\disp_x^2 +1 )^{1/2} dx /\int_\D dx $, respectively, while Fig. \ref{fig:err_MS} portrays accumulated model errors through recurrent prediction steps.
To quantify the statistics of long-term predicted solutions, an auto-correlation function is introduced:
\begin{equation}
\mathcal{R} (r)   = \mathbb{E} \left(
 \int_\D \phi^*(x) \phi^*(x-r) dx 
 /\int_\D \phi^*(x) \phi^*(x) dx 
 \right).
\label{eq:corr}
\end{equation}
where $\phi^*(x)$ denotes a predicted solution obtained after a sufficiently long time. Fig. \ref{fig:corr_MS} compares the auto-correlation function obtained by all models with the reference one.
For learning the other operator $\hat{\mathcal{G}}_\beta^{DT}$, similar plots are presented for front displacement (Fig. \ref{fig:disp_KS}), front slope  (Fig. \ref{fig:uSlope_KS}), total flame front length (Fig. \ref{fig:len_KS}), accumulated error (Fig. \ref{fig:err_KS}), and auto-correlation (Fig. \ref{fig:corr_KS}).

The findings can be summarized as follows:

\begin{enumerate}
\item 
 All three models-pFNO, pFNO*, and pCNN-perform generally well in learning two parametric front evolution operators of $\hat{\mathcal{G}}_\nu^{DL}$ and $\hat{\mathcal{G}}_\beta^{DT}$.  

When learning the DL-destabilized front evolution at varying parameter values $\nu$, all models demonstrate proficiency in making reasonably accurate short-term predictions ($t/\Delta_t \leq 50$). This is evident from the small relative errors (less than $0.02$ as shown in Table \ref{Table1} and Fig. \ref{fig:err_MS}), as well as from the predicted front displacements, front slope, and total front length illustrated in Figs. \ref{fig:disp_MS}, \ref{fig:uSlope_MS}, and \ref{fig:len_MS}, respectively. The long-term predictions ($t/\Delta_t \geq 500$) by all models align with the characteristic reference solution pattern, specifically the persistent noise-affected single-cusp fronts depicted in Figs. \ref{fig:disp_MS} and \ref{fig:uSlope_MS}, and approximate the auto-correlation functions shown in Fig. \ref{fig:corr_MS}.

Similarly, on learning the evolution of DT-fronts at different parameter values $\beta$, all three models excel in predicting short-term solutions (indicated by the errors shown in Table \ref{Table1} and Fig. \ref{fig:err_KS}, and by the results over $t/\Delta_t \leq 50$ in Figs \ref{fig:disp_KS}, \ref{fig:uSlope_KS}, \ref{fig:len_KS}). Furthermore, they successfully capture the statistical characteristics of long-term chaotic solutions, as evidenced by the auto-correlations shown in Fig. \ref{fig:corr_MS}.

\item

 In term of deciphering the prolonged evolution of chaotic DT-fronts, pFNO outperforms the other two models, pFNO* and pCNN, as demonstrated by the superior auto-correlations at five distinct $\beta$ values illustrated in Fig. \ref{fig:corr_KS}. Moreover, pFNO provides the most accurate short-term predictions for both DT-fronts and DL-fronts, supported by the smallest errors highlighted in Table \ref{Table1}, and depicted in Figs. \ref{fig:err_MS} and \ref{fig:err_KS}.

However, when faced with the challenge of learning the long-term evolution of DL-fronts, whose characteristics depend on the parameter value $\nu$ and are prone to noise, pFNO is outperformed by pFNO* and pCNN. This is evident in the comparison of auto-correlation functions shown in Fig. \ref{fig:corr_MS}. In particular, all three models tend to overestimate the impact of noise-induced front wrinkles at larger values of $\nu \geq 0.07$, as observed in Figs. \ref{fig:disp_MS} and \ref{fig:uSlope_MS}. Here, the predicted front displacement and slope remain frequently disturbed, while the reference solutions evolve to become steady at $\nu=0.15$. Consequently, all models also overpredict the total front length at large $\nu$, as shown in Fig. \ref{fig:len_MS}.

Moreover, Fig. \ref{fig:len_MS} reveals that pFNO* and pFNO tend to predict a higher total front length than pCNN, especially at small parameter values of $\nu \leq 0.035$. This behavior may be attributed to the Fourier operator-based method's tendency to fit noise signals through certain high-frequency representations.

It is noteworthy that in the previous study\cite{Yu2023}, while the (non-parametric) baseline CNN method has demonstrated the ability to learn and reproduce the reference steady DL-front solution at large $\nu$, when trained separately to learn the solutions at small $\nu$, it tends to predict significant artifacts. Interestingly, our proposed parametric-CNN model, represented by a single network, now yields good predictions for the solution at small $\nu$, albeit with less accuracy in predicting the steady solution at large $\nu$.

\end{enumerate}

\subsection{2d training dataset}

The 2D training dataset utilized in this study is derived from DNS presented in a previous work \cite{YBB15PRE}. These simulations focus on the nonlinear evolution of an unstable premixed flame due to the DL stability. The governing system comprises reactive Navier-Stokes equations with a low Mach number approximation, assuming a one-step Arrhenius reaction and a unity Lewis number \cite{YBB15PRE}. The DL stability is maintained by sustaining an above-unity density ratio ($\densityRatio$) between the cold fresh gas and hot burned gas on either side of the flame front. High-order numerical methods \cite{YYB12, YB14, YB13Numeric} are employed for computational solutions. The study explores the free propagation of an initially planar flame into a quiescent fresh reactant within a 2D periodic channel of width $\channelWidth$. The computational domain is a rectangular window region $ [0, \channelWidth] \times [0, 3\channelWidth]$, moving with the propagating flame \cite{YBB15PRE, YU2017nonlinear, YL2019_Equation, YNBL2019_CF,  YNCL2020_JFM}.

In the aforementioned DNS study \cite{YBB15PRE}, comprehensive simulations were conducted covering relevant parameter ranges for different channel widths ($\channelWidth$) and various flames characterized by the density ratio ($\densityRatio$). Each DNS is a single-instance run commencing from an initially perturbed planar flame, and the simulation extends to a sufficiently long time to allow for the nonlinear development of flame instability. Each DNS run outputs a sequence of 2D flame fronts.

Here, a `2D flame front' of $\partial \Omega$ refers to the boundary of the `burned' domain $ \Omega(t) =\{x | Y(x;t)> 0.6 \}$, with $Y$ is the fuel mass fraction being 0 in the fresh reactant and 1 in the product.
 In other words, 2D flame fronts are a collection of iso-scalar lines of $Y(x,t)=0.6$ numerically extracted from a $x_j$-mesh representation of $Y(x_j,t)$ using the marching square method.

For the operator learning methods to describe the 2D flame evolution, the iso-lines representation of zero-thickness flame fronts is extended to a functional representation of $\atanG(x,t)$:
\begin{equation}
\begin{split}
\atanG(x,t) &= 
\text{tanh}\left(  \frac{ \varsigma }{\Delta_*} \distanceFunc \left(x,\partial \Omega(t) \right)   \right)  
\label{eq:G_atan}
\end{split}
\end{equation}
Here, tanh represents the hyperbolic tangent, $\Delta_{*}$ is a factor allowing the front to have a finite thickness, and $\varsigma$ = 1 if $x\in \Omega$ and -1 otherwise. For any point $x\in \D \subset \mathbb{R}^2$ (with $\int_{\D} Y(x,t) dx = 0$), the function $\distanceFunc (x,\partial \Omega) := \inf_{ y \in \partial \Omega \cap y_1=x_1} |x_2-y_2|$ measures the smallest distance to the points in $\partial \Omega$ sharing the same first coordinate of $x_1$.

In this study, we utilize training data extracted from four representative DNS simulations of the same flame characterized by a density ratio ($\densityRatio=8$) but in different channel widths: $\channelWidth/\flamethickness=[320, 512, 768, 1536]$, where $\flamethickness$ represents the laminar flame thickness. The final DNS simulation focuses on flame development in an exceptionally wide channel, resulting in fractal-like, complex flame front structures. It is worth highlighting that obtaining these DNS results required approximately 2 million CPU hours.

To better understand the influence of 2D flame instability caused by variations in channel width, we employ a parametric ratio $\mu = \cutoffWidth/\channelWidth$, where $\cutoffWidth$ is the channel width below which the investigated flame remains stable. This value is determined through numerical testing to be $\cutoffWidth = 17 \flamethickness$. The four DNS cases are characterized by $\mu \in [0.053, 0.033, 0.022, 0.011]$, with $\mu$ being comparable to $\nu$ in the 1D MS Equation \eqref{eq:MS}.

We choose a sufficiently large domain $\D=[0,\channelWidth ] \times[-0.45\channelWidth, 1.15\channelWidth]$ to encompass all flame elements obtained from these DNS runs. The 2D operator-learning networks take inputs as a discrete representation of the function $\atanG(x_j,t)$, evaluated on a uniform mesh $x_j$ discretizing $\D$ with a mesh size of $256\times 256$. All cases adopt the same thickness factor $\Delta_{*}= \channelWidth/128$. 
Illustrations of the functional representation of $\atanG(x)$ can be seen in the 2D-pCNN depicted in Figure \ref{fig:CNN}, where its input and output functions $v(x)$ and $v'(x)$ are shown. Here, $\atanG(x)$ takes values of either -1 (yellow color) or +1 (blue color) due to the tanh function in Eq. \eqref{eq:G_atan}, and can also take intermediate values near the region of zero iso-lines of $\atanG(x,t)=0$.

The four DNS simulations extend for a duration of $[279, 269, 343, 181]\cutoffWidth/\flamespeed$, where $\flamespeed$ denotes the laminar flame speed. Each DNS outputs a consecutive sequence of 2D flame fronts $\partial \Omega$ at a time interval of $0.26 \cutoffWidth/\flamespeed$, resulting in a total number of $N_\mu = [1075, 1047, 1335, 703]$ consecutive 2D flame fronts. Samples of DNS fronts are shown in the first row of Figure \ref{fig:disp2d}, highlighting the complexity of DNS fronts, which cannot be adequately represented as 1D functions.

The model training relies on pairing the four DNS front sequences in a 1-to-$n$ fashion, specifically $\{ \atanG(t_{j}), [\atanG(t_{j+1}),...,\atanG(t_{j+n})] \}$, where $t_k=k\Delta_t$ and $j=j' \cdot m_\mu $ for $j'= 1,...,\frac{N_\mu - n}{m_\mu}$, and $m_\mu =[2,2,2,1]$ respective to four DNS runs with $\mu$ in decreasing order. Due to the limited dataset size, no data is reserved for validation. Additionally, constrained by computer hardware memory, we set $n$= 10 for training both 2D-pFNO and 2D-pCNN models.

\subsection{Learning 2d flame evolution}

The relative $L^2$ training errors for pFNO and pCNN are 0.0094 and 0.0055, respectively, indicating proficient learning in both methods.

\begin{figure*}
	\centerline{
		\includegraphics[width=\linewidth]{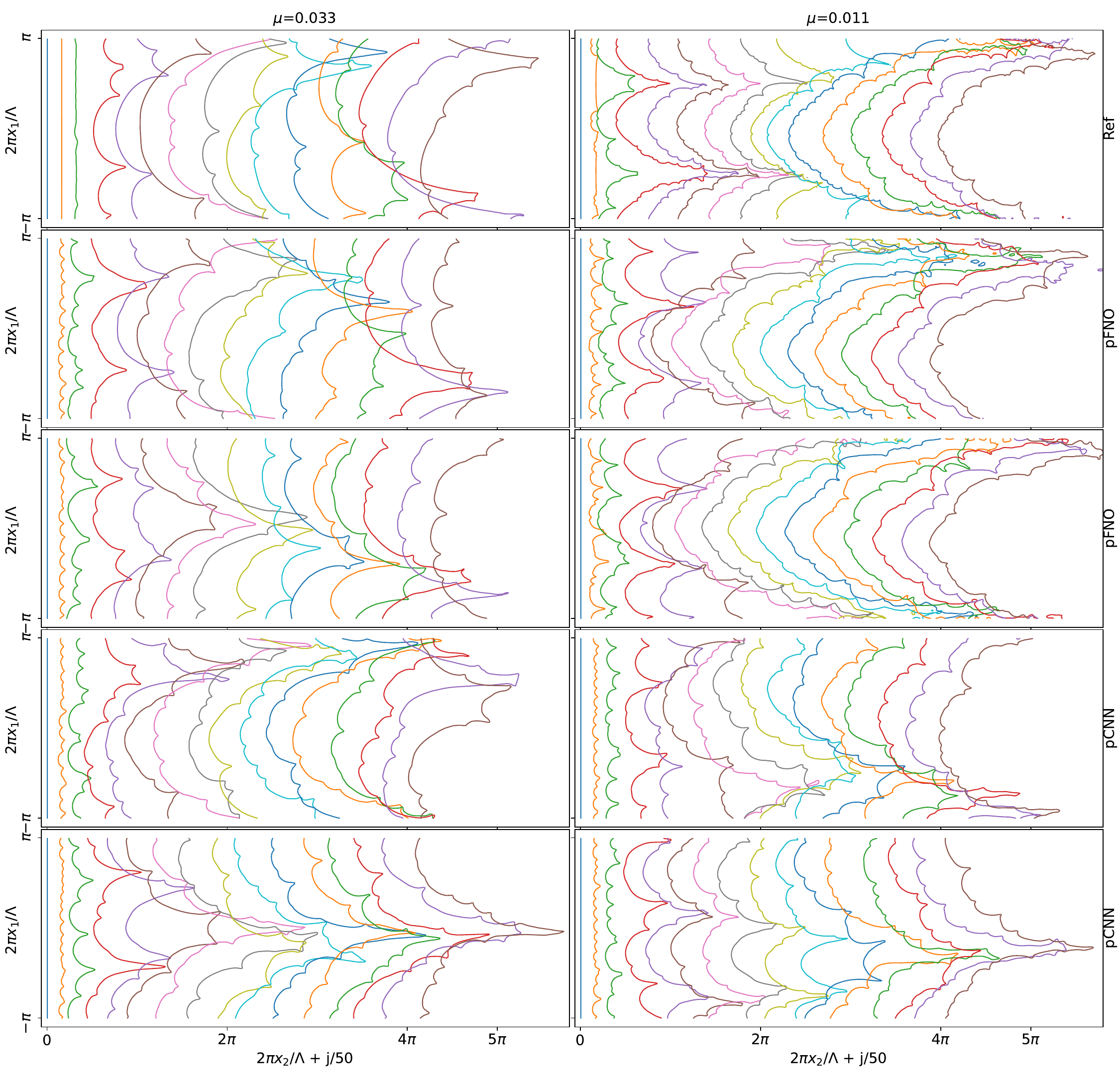}   
}
	\caption{
		\label{fig:disp2d}
Comparison of front sequences, $\partial \Omega(t_j)$, originating from two DNS runs (top row) capturing the unstable flame evolution in channels characterized by $\mu$ values of 0.033 (left column) and 0.011 (right column). These ground truth sequences are juxtaposed with predictions generated by the 2D-pFNO model (second and third rows) and the 2D-pCNN model (the last two rows). Each sequence commences from a random initial condition, and snapshots at selected time instances $t_j = j \Delta_t$ for $j=[0, 25, 50, 100, 150, 200, 250, 300, 350, 400, 450, 500, 550, 600, 650, 700]$ are sequentially displayed from left to right.
	}
\end{figure*}

Figure \ref{fig:disp2d} illustrates the flame fronts extracted from two reference DNS cases with $\nu$ values of 0.033 and 0.011, comparing them to the predictions generated by two parametric operator learning models, 2D-pFNO and 2D-pCNN. Both models successfully capture the general trend of $\nu$-dependent front evolution. In the case of flame development in a larger channel (characterized by $\mu=0.011$), both models replicate long-term intricate front structures, featuring frequent noisy wrinkles atop cellular shapes of varying sizes. Conversely, for the smaller channel with $\nu=0.033$, the models predict a considerably smoother front evolution, albeit with a slight overestimation of noisy wrinkling-a phenomenon observed similarly when learning 1D DL-fronts.

Given that the models have learned the underlying flame front evolution operators, they demonstrate the ability to predict new instance solutions starting from random initial conditions, as depicted in Figure \ref{fig:disp2d}.

\section{Summary and Conclusion \label{sec:conclusion} }

In this study, we delve into the application of parametric learning methods, specifically Fourier Neural Operator (FNO) and Convolutional Neural Network (CNN), to forecast the evolution of unstable flame fronts in both one and two dimensions. These methods are extended to assimilate additional input parameters, enriching our ability to comprehend the nuanced and varied conditions governing flame evolution.

The proposed methods, namely pCNN and pFNO, are applied to model the flame front evolution described by two 1D parametric partial differential equations: the Michael-Sivashinsky (MS) and Kuramoto-Sivashinsky (KS) equations, corresponding to the Darrieus-Landau (DL) and diffusive-Thermal (DT) instabilities, respectively. Across the relevant parameter range, both models exhibit proficiency in accurate short-term predictions for both PDEs. When learning solutions of the KS equation, both models adeptly capture the statistical characteristics of long-term chaotic solutions, faithfully reproducing auto-correlation functions comparable to reference solutions.

In the context of long-term solutions to the MS equations, both models capture the peculiar phenomenon of parameter-dependent increasing noise disrupting a stationary solution. However, they tend to overestimate noise-induced unresting front wrinkles when the reference solution maintains a stable state against noise disturbance. Notably, when the reference long-term solutions exhibit highly disturbed noisy patterns, pCNN outperforms pFNO in predicting front length accurately.

Expanding our study to 2D flame evolution, we leverage direct numerical simulations (DNS) for training data. Our proposed parametric learning models, 2D-pFNO and 2D-pCNN, adeptly capture the channel width-dependent front evolution. In larger channels, both models reproduce intricate long-term front structures, while in smaller channels, they predict a smoother evolution with a slight overestimation of noisy wrinkling. Both 2D models successfully learn the underlying flame front evolution operators, enabling predictions of new flame solutions from random initial conditions.

This work highlights the effectiveness of parametric learning methods in predicting the evolution of unstable flame fronts. The models excel in capturing short-term dynamics and replicating long-term statistical characteristics. However, challenges emerge in learning the long-term evolution of DL-fronts, particularly in the presence of noise. The study offers valuable insights into the capabilities and limitations of parametric learning methods in the realm of flame dynamics, paving the way for further advancements in predicting complex physical phenomena.

\begin{acknowledgments}
The author gratefully acknowledges the financial support by the Swedish Research Council (VR-2019-05648). 
The simulations were performed using the computer facilities provided by the Swedish National Infrastructure for Computing (SNIC) at PDC , HPC2N and ALVIS.
This work benefited from the preliminary study conducted during Ludvig Nobel's master's thesis at Lund University in 2022.
\end{acknowledgments}

\section*{Data Availability Statement}
The code and data that support the findings of this study are openly available at www.github.com/RixinYu/ML\_paraFlame.

\section*{Declaration of Interest}
The authors have no conflicts to disclose.


\appendix
\section{ Network hyper-parameters and training details  \label{app:nn_detail} }

All models are trained for 1000 epochs using the Adam optimizer with a learning rate of 0.0025 and weight decay of 0.0001. A scheduler step size of 100 and gamma of 0.5 are employed for learning rate adjustment.
For all 1D networks trained in a 1-to-20 manner, a batch size of 800 is utilized. During model training, the maximum norm of gradients is clipped above 50 to stabilize the training process.
The 2D-pFNO networks are trained in a 1-to-10 manner with a batch size of 40, while the 2D-pCNN networks are trained similarly in a 1-to-10 manner with a batch size of 28. These batch sizes are determined by the available GPU memory (NVIDIA Tesla A40).
All trainings are performed using a single GPU, training a  1D-pFNO model takes around 20 hours, while training a 1D-pCNN takes 40 hours. Training the 2D-pFNO and 2D-pCNN models takes around 52 hours and 48 hours respectively.  

All 1D-pFNO networks utilize $L=4$ levels of parametric Fourier layers and $d_{\z}=30$ channels. For learning the MS equation, two hyperparameters, $\kk_\text{max}=64$ and $N_\gamma=6$, are employed, while for the KS equation, these are slightly adjusted to $\kk_\text{max}=128$ and $N_\gamma=5$.
The 2D-pFNO network adopts hyperparameters with $L=4$, $d_{\z}=20$, $\kk_{(1),\text{max}}=\kk_{(2),\text{max}}=64$, and $N_\gamma=6$.
All pFNO methods are implemented by sharing most of the trainable parameters within a single Fourier layer (Eq. \eqref{eq:pFNO}) across all layers $l = 0, ..., L-1$, except for those used to parameterize the function $D_l(\gamma)$. This adjustment significantly reduces the model size with minimal impact on performance, making it a standard practice across all pFNO models. Additionally, a skip connection may be added to the Fourier layer as
\begin{equation}
\z_{l+1} = \z_{l} + \sigma \left( \F^{-1} \{ \mathfrak{R}^*_l ( \F \{ \z_l \} , \gamma ) \} + W_l (\z_l) \right).
\label{eq:FNO_skip}
\end{equation}
Adding such a skip connection has been found to facilitate model training for 2D-pFNO.

The 1D-pCNN shares a similar architecture with the 2D-pCNN illustrated in Figure \ref{fig:CNN} except a few difference:
(i) All images ($e$, $e^*$, $e^+$, and $e'$) become 1D, necessitating the adaptation of convolution layers, up-sample layers, and max-pooling layers to their corresponding 1D versions.
(ii) The parameter influence is activated only for the first 4 encoder levels, effectively setting the function $D_l=0$ for $l\geq4$.
(iii) The channel number after the first encoder layer ($c_1$) changes to $16$.

\bibliography{ReferencesML}

\begin{thebibliography}{47}%
\makeatletter
\providecommand \@ifxundefined [1]{%
 \@ifx{#1\undefined}
}%
\providecommand \@ifnum [1]{%
 \ifnum #1\expandafter \@firstoftwo
 \else \expandafter \@secondoftwo
 \fi
}%
\providecommand \@ifx [1]{%
 \ifx #1\expandafter \@firstoftwo
 \else \expandafter \@secondoftwo
 \fi
}%
\providecommand \natexlab [1]{#1}%
\providecommand \enquote  [1]{``#1''}%
\providecommand \bibnamefont  [1]{#1}%
\providecommand \bibfnamefont [1]{#1}%
\providecommand \citenamefont [1]{#1}%
\providecommand \href@noop [0]{\@secondoftwo}%
\providecommand \href [0]{\begingroup \@sanitize@url \@href}%
\providecommand \@href[1]{\@@startlink{#1}\@@href}%
\providecommand \@@href[1]{\endgroup#1\@@endlink}%
\providecommand \@sanitize@url [0]{\catcode `\\12\catcode `\$12\catcode
  `\&12\catcode `\#12\catcode `\^12\catcode `\_12\catcode `\%12\relax}%
\providecommand \@@startlink[1]{}%
\providecommand \@@endlink[0]{}%
\providecommand \url  [0]{\begingroup\@sanitize@url \@url }%
\providecommand \@url [1]{\endgroup\@href {#1}{\urlprefix }}%
\providecommand \urlprefix  [0]{URL }%
\providecommand \Eprint [0]{\href }%
\providecommand \doibase [0]{https://doi.org/}%
\providecommand \selectlanguage [0]{\@gobble}%
\providecommand \bibinfo  [0]{\@secondoftwo}%
\providecommand \bibfield  [0]{\@secondoftwo}%
\providecommand \translation [1]{[#1]}%
\providecommand \BibitemOpen [0]{}%
\providecommand \bibitemStop [0]{}%
\providecommand \bibitemNoStop [0]{.\EOS\space}%
\providecommand \EOS [0]{\spacefactor3000\relax}%
\providecommand \BibitemShut  [1]{\csname bibitem#1\endcsname}%
\let\auto@bib@innerbib\@empty
\bibitem [{\citenamefont {Guo}, \citenamefont {Li},\ and\ \citenamefont
  {Iorio}(2016)}]{CNN1}%
  \BibitemOpen
  \bibfield  {author} {\bibinfo {author} {\bibfnamefont {X.}~\bibnamefont
  {Guo}}, \bibinfo {author} {\bibfnamefont {W.}~\bibnamefont {Li}},\ and\
  \bibinfo {author} {\bibfnamefont {F.}~\bibnamefont {Iorio}},\ }\bibfield
  {title} {\enquote {\bibinfo {title} {Convolutional neural networks for steady
  ﬂow approximation},}\ }in\ \href@noop {} {\emph {\bibinfo {booktitle}
  {Proceedings of the 22nd ACM SIGKDD International Conference on Knowledge
  Discovery and Data Mining}}}\ (\bibinfo {year} {2016})\BibitemShut {NoStop}%
\bibitem [{\citenamefont {Zhu}\ and\ \citenamefont {Zabaras}(2018)}]{CNN2}%
  \BibitemOpen
  \bibfield  {author} {\bibinfo {author} {\bibfnamefont {Y.}~\bibnamefont
  {Zhu}}\ and\ \bibinfo {author} {\bibfnamefont {N.}~\bibnamefont {Zabaras}},\
  }\bibfield  {title} {\enquote {\bibinfo {title} {Bayesian deep convolutional
  encoder–decoder networks for surrogate modeling and uncertainty
  quantification},}\ }\href@noop {} {\bibfield  {journal} {\bibinfo  {journal}
  {Journal of Computational Physics}\ }\textbf {\bibinfo {volume} {366}},\
  \bibinfo {pages} {415--447} (\bibinfo {year} {2018})}\BibitemShut {NoStop}%
\bibitem [{\citenamefont {Adler}\ and\ \citenamefont {Öktem}(2017)}]{CNN3}%
  \BibitemOpen
  \bibfield  {author} {\bibinfo {author} {\bibfnamefont {J.}~\bibnamefont
  {Adler}}\ and\ \bibinfo {author} {\bibfnamefont {O.}~\bibnamefont {Öktem}},\
  }\bibfield  {title} {\enquote {\bibinfo {title} {Solving ill-posed inverse
  problems using iterative deep neural networks},}\ }\href@noop {} {\bibfield
  {journal} {\bibinfo  {journal} {Inverse Problems}\ }\textbf {\bibinfo
  {volume} {33}},\ \bibinfo {pages} {124007} (\bibinfo {year}
  {2017})}\BibitemShut {NoStop}%
\bibitem [{\citenamefont {Bhatnagar}\ \emph {et~al.}(2019)\citenamefont
  {Bhatnagar}, \citenamefont {Afshar}, \citenamefont {Pan}, \citenamefont
  {Duraisamy},\ and\ \citenamefont {Kaushik}}]{CNN4}%
  \BibitemOpen
  \bibfield  {author} {\bibinfo {author} {\bibfnamefont {S.}~\bibnamefont
  {Bhatnagar}}, \bibinfo {author} {\bibfnamefont {Y.}~\bibnamefont {Afshar}},
  \bibinfo {author} {\bibfnamefont {S.}~\bibnamefont {Pan}}, \bibinfo {author}
  {\bibfnamefont {K.}~\bibnamefont {Duraisamy}},\ and\ \bibinfo {author}
  {\bibfnamefont {S.}~\bibnamefont {Kaushik}},\ }\bibfield  {title} {\enquote
  {\bibinfo {title} {Prediction of aerodynamic flow fields using convolutional
  neural networks},}\ }\href@noop {} {\bibfield  {journal} {\bibinfo  {journal}
  {Computational Mechanics}\ }\textbf {\bibinfo {volume} {64}},\ \bibinfo
  {pages} {525--545} (\bibinfo {year} {2019})}\BibitemShut {NoStop}%
\bibitem [{\citenamefont {KHOO}, \citenamefont {LU},\ and\ \citenamefont
  {YING}(2021)}]{CNN5}%
  \BibitemOpen
  \bibfield  {author} {\bibinfo {author} {\bibfnamefont {Y.}~\bibnamefont
  {KHOO}}, \bibinfo {author} {\bibfnamefont {J.}~\bibnamefont {LU}},\ and\
  \bibinfo {author} {\bibfnamefont {L.}~\bibnamefont {YING}},\ }\bibfield
  {title} {\enquote {\bibinfo {title} {Solving parametric pde problems with
  artificial neural networks},}\ }\href
  {https://doi.org/10.1017/S0956792520000182} {\bibfield  {journal} {\bibinfo
  {journal} {European Journal of Applied Mathematics}\ }\textbf {\bibinfo
  {volume} {32}},\ \bibinfo {pages} {421–435} (\bibinfo {year}
  {2021})}\BibitemShut {NoStop}%
\bibitem [{\citenamefont {Ronneberger}, \citenamefont {Fischer},\ and\
  \citenamefont {Brox}(2015)}]{UNet}%
  \BibitemOpen
  \bibfield  {author} {\bibinfo {author} {\bibfnamefont {O.}~\bibnamefont
  {Ronneberger}}, \bibinfo {author} {\bibfnamefont {P.}~\bibnamefont
  {Fischer}},\ and\ \bibinfo {author} {\bibfnamefont {T.}~\bibnamefont
  {Brox}},\ }\href@noop {} {\enquote {\bibinfo {title} {U-net: Convolutional
  networks for biomedical image segmentation},}\ } (\bibinfo {year} {2015}),\
  \Eprint {https://arxiv.org/abs/1505.04597} {arXiv:1505.04597 [cs.CV]}
  \BibitemShut {NoStop}%
\bibitem [{\citenamefont {Winovich}, \citenamefont {Ramani},\ and\
  \citenamefont {Lin}(2019)}]{ConvPDE}%
  \BibitemOpen
  \bibfield  {author} {\bibinfo {author} {\bibfnamefont {N.}~\bibnamefont
  {Winovich}}, \bibinfo {author} {\bibfnamefont {K.}~\bibnamefont {Ramani}},\
  and\ \bibinfo {author} {\bibfnamefont {G.}~\bibnamefont {Lin}},\ }\bibfield
  {title} {\enquote {\bibinfo {title} {Convpde-uq: Convolutional neural
  networks with quantified uncertainty for heterogeneous elliptic partial
  differential equations on varied domains},}\ }\href
  {https://doi.org/https://doi.org/10.1016/j.jcp.2019.05.026} {\bibfield
  {journal} {\bibinfo  {journal} {Journal of Computational Physics}\ }\textbf
  {\bibinfo {volume} {394}},\ \bibinfo {pages} {263--279} (\bibinfo {year}
  {2019})}\BibitemShut {NoStop}%
\bibitem [{\citenamefont {Li}\ \emph {et~al.}(2020{\natexlab{a}})\citenamefont
  {Li}, \citenamefont {Kovachki}, \citenamefont {Azizzadenesheli},
  \citenamefont {Liu}, \citenamefont {Bhattacharya}, \citenamefont {Stuart},\
  and\ \citenamefont {Anandkumar}}]{GraphKenerlNetwork}%
  \BibitemOpen
  \bibfield  {author} {\bibinfo {author} {\bibfnamefont {Z.}~\bibnamefont
  {Li}}, \bibinfo {author} {\bibfnamefont {N.}~\bibnamefont {Kovachki}},
  \bibinfo {author} {\bibfnamefont {K.}~\bibnamefont {Azizzadenesheli}},
  \bibinfo {author} {\bibfnamefont {B.}~\bibnamefont {Liu}}, \bibinfo {author}
  {\bibfnamefont {K.}~\bibnamefont {Bhattacharya}}, \bibinfo {author}
  {\bibfnamefont {A.}~\bibnamefont {Stuart}},\ and\ \bibinfo {author}
  {\bibfnamefont {A.}~\bibnamefont {Anandkumar}},\ }\href@noop {} {\enquote
  {\bibinfo {title} {Neural operator: Graph kernel network for partial
  differential equations},}\ } (\bibinfo {year} {2020}{\natexlab{a}}),\ \Eprint
  {https://arxiv.org/abs/2003.03485} {arXiv:2003.03485 [cs.LG]} \BibitemShut
  {NoStop}%
\bibitem [{\citenamefont {Kovachki}\ \emph {et~al.}(2021)\citenamefont
  {Kovachki}, \citenamefont {Li}, \citenamefont {Liu}, \citenamefont
  {Azizzadenesheli}, \citenamefont {Bhattacharya}, \citenamefont {Stuart},\
  and\ \citenamefont {Anandkumar}}]{kovachki2021neural}%
  \BibitemOpen
  \bibfield  {author} {\bibinfo {author} {\bibfnamefont {N.}~\bibnamefont
  {Kovachki}}, \bibinfo {author} {\bibfnamefont {Z.}~\bibnamefont {Li}},
  \bibinfo {author} {\bibfnamefont {B.}~\bibnamefont {Liu}}, \bibinfo {author}
  {\bibfnamefont {K.}~\bibnamefont {Azizzadenesheli}}, \bibinfo {author}
  {\bibfnamefont {K.}~\bibnamefont {Bhattacharya}}, \bibinfo {author}
  {\bibfnamefont {A.}~\bibnamefont {Stuart}},\ and\ \bibinfo {author}
  {\bibfnamefont {A.}~\bibnamefont {Anandkumar}},\ }\bibfield  {title}
  {\enquote {\bibinfo {title} {Neural operator: Learning maps between function
  spaces},}\ }\href@noop {} {\bibfield  {journal} {\bibinfo  {journal} {arXiv
  preprint arXiv:2108.08481}\ } (\bibinfo {year} {2021})}\BibitemShut {NoStop}%
\bibitem [{\citenamefont {Lu}, \citenamefont {Jin},\ and\ \citenamefont
  {Karniadakis}(2019)}]{DeepONet}%
  \BibitemOpen
  \bibfield  {author} {\bibinfo {author} {\bibfnamefont {L.}~\bibnamefont
  {Lu}}, \bibinfo {author} {\bibfnamefont {P.}~\bibnamefont {Jin}},\ and\
  \bibinfo {author} {\bibfnamefont {G.}~\bibnamefont {Karniadakis}},\
  }\href@noop {} {\enquote {\bibinfo {title} {Deeponet: Learning nonlinear
  operators for identifying differential equations based on the universal
  approximation theorem of operators},}\ } (\bibinfo {year} {2019}),\ \Eprint
  {https://arxiv.org/abs/1910.03193} {arXiv:1910.03193 [cs.LG]} \BibitemShut
  {NoStop}%
\bibitem [{\citenamefont {Li}\ \emph {et~al.}(2020{\natexlab{b}})\citenamefont
  {Li}, \citenamefont {Kovachki}, \citenamefont {Azizzadenesheli},
  \citenamefont {Liu}, \citenamefont {Bhattacharya}, \citenamefont {Stuart},\
  and\ \citenamefont {Anandkumar}}]{FNO2020}%
  \BibitemOpen
  \bibfield  {author} {\bibinfo {author} {\bibfnamefont {Z.}~\bibnamefont
  {Li}}, \bibinfo {author} {\bibfnamefont {N.}~\bibnamefont {Kovachki}},
  \bibinfo {author} {\bibfnamefont {K.}~\bibnamefont {Azizzadenesheli}},
  \bibinfo {author} {\bibfnamefont {B.}~\bibnamefont {Liu}}, \bibinfo {author}
  {\bibfnamefont {K.}~\bibnamefont {Bhattacharya}}, \bibinfo {author}
  {\bibfnamefont {A.}~\bibnamefont {Stuart}},\ and\ \bibinfo {author}
  {\bibfnamefont {A.}~\bibnamefont {Anandkumar}},\ }\href@noop {} {\enquote
  {\bibinfo {title} {Fourier neural operator for parametric partial
  differential equations},}\ } (\bibinfo {year} {2020}{\natexlab{b}}),\ \Eprint
  {https://arxiv.org/abs/2010.08895} {arXiv:2010.08895 [cs.LG]} \BibitemShut
  {NoStop}%
\bibitem [{\citenamefont {N.}, \citenamefont {S.},\ and\ \citenamefont
  {S.}(2021)}]{FNO_theroy}%
  \BibitemOpen
  \bibfield  {author} {\bibinfo {author} {\bibfnamefont {K.}~\bibnamefont
  {N.}}, \bibinfo {author} {\bibfnamefont {L.}~\bibnamefont {S.}},\ and\
  \bibinfo {author} {\bibfnamefont {M.}~\bibnamefont {S.}},\ }\href@noop {}
  {\enquote {\bibinfo {title} {On universal approximation and error bounds for
  fourier neural operators},}\ } (\bibinfo {year} {2021}),\ \Eprint
  {https://arxiv.org/abs/2107.07562} {arXiv:2107.07562 [math.NA]} \BibitemShut
  {NoStop}%
\bibitem [{\citenamefont {T.}\ and\ \citenamefont
  {H.}(1995)}]{DeepONet_ChenChen_Theory}%
  \BibitemOpen
  \bibfield  {author} {\bibinfo {author} {\bibfnamefont {C.}~\bibnamefont
  {T.}}\ and\ \bibinfo {author} {\bibfnamefont {C.}~\bibnamefont {H.}},\
  }\bibfield  {title} {\enquote {\bibinfo {title} {Universal approximation to
  nonlinear operators by neural networks with arbitrary activation functions
  and its application to dynamical systems},}\ }\href@noop {} {\bibfield
  {journal} {\bibinfo  {journal} {IEEE Trans. Neural Netw.}\ }\textbf {\bibinfo
  {volume} {6}},\ \bibinfo {pages} {911--917} (\bibinfo {year}
  {1995})}\BibitemShut {NoStop}%
\bibitem [{\citenamefont {Lanthaler}, \citenamefont {Li},\ and\ \citenamefont
  {Stuart}(2023)}]{lanthaler2023nonlocal}%
  \BibitemOpen
  \bibfield  {author} {\bibinfo {author} {\bibfnamefont {S.}~\bibnamefont
  {Lanthaler}}, \bibinfo {author} {\bibfnamefont {Z.}~\bibnamefont {Li}},\ and\
  \bibinfo {author} {\bibfnamefont {A.~M.}\ \bibnamefont {Stuart}},\ }\bibfield
   {title} {\enquote {\bibinfo {title} {The nonlocal neural operator: Universal
  approximation},}\ }\href@noop {} {\bibfield  {journal} {\bibinfo  {journal}
  {arXiv preprint arXiv:2304.13221}\ } (\bibinfo {year} {2023})}\BibitemShut
  {NoStop}%
\bibitem [{\citenamefont {Lu}\ \emph {et~al.}(2021)\citenamefont {Lu},
  \citenamefont {Jin}, \citenamefont {Pang}, \citenamefont {Zhang},\ and\
  \citenamefont {Karniadakis}}]{DeepONet_Nature}%
  \BibitemOpen
  \bibfield  {author} {\bibinfo {author} {\bibfnamefont {L.}~\bibnamefont
  {Lu}}, \bibinfo {author} {\bibfnamefont {P.}~\bibnamefont {Jin}}, \bibinfo
  {author} {\bibfnamefont {G.}~\bibnamefont {Pang}}, \bibinfo {author}
  {\bibfnamefont {Z.}~\bibnamefont {Zhang}},\ and\ \bibinfo {author}
  {\bibfnamefont {G.}~\bibnamefont {Karniadakis}},\ }\bibfield  {title}
  {\enquote {\bibinfo {title} {Learning nonlinear operators via deeponet based
  on the universal approximation theorem of operators},}\ }\href
  {https://doi.org/https://doi.org/10.1038/s42256-021-00302-5} {\bibfield
  {journal} {\bibinfo  {journal} {Nat. Mach. Intell.}\ }\textbf {\bibinfo
  {volume} {3}},\ \bibinfo {pages} {218--229} (\bibinfo {year}
  {2021})}\BibitemShut {NoStop}%
\bibitem [{\citenamefont {Lu}\ \emph {et~al.}(2022)\citenamefont {Lu},
  \citenamefont {Meng}, \citenamefont {Cai}, \citenamefont {Mao}, \citenamefont
  {Goswami}, \citenamefont {Zhang},\ and\ \citenamefont
  {Karniadakis}}]{DeepONet_FNO_cmp}%
  \BibitemOpen
  \bibfield  {author} {\bibinfo {author} {\bibfnamefont {L.}~\bibnamefont
  {Lu}}, \bibinfo {author} {\bibfnamefont {X.}~\bibnamefont {Meng}}, \bibinfo
  {author} {\bibfnamefont {S.}~\bibnamefont {Cai}}, \bibinfo {author}
  {\bibfnamefont {Z.}~\bibnamefont {Mao}}, \bibinfo {author} {\bibfnamefont
  {S.}~\bibnamefont {Goswami}}, \bibinfo {author} {\bibfnamefont
  {Z.}~\bibnamefont {Zhang}},\ and\ \bibinfo {author} {\bibfnamefont
  {G.}~\bibnamefont {Karniadakis}},\ }\bibfield  {title} {\enquote {\bibinfo
  {title} {A comprehensive and fair comparison of two neural operators (with
  practical extensions) based on fair data},}\ }\href
  {https://doi.org/https://doi.org/10.1038/s42256-021-00302-5} {\bibfield
  {journal} {\bibinfo  {journal} {Computer Methods in Applied Mechanics and
  Engineering}\ }\textbf {\bibinfo {volume} {393}},\ \bibinfo {pages} {114778}
  (\bibinfo {year} {2022})}\BibitemShut {NoStop}%
\bibitem [{\citenamefont {Gupta}, \citenamefont {Xiao},\ and\ \citenamefont
  {Bogdan}(2021)}]{gupta2021multiwavelet}%
  \BibitemOpen
  \bibfield  {author} {\bibinfo {author} {\bibfnamefont {G.}~\bibnamefont
  {Gupta}}, \bibinfo {author} {\bibfnamefont {X.}~\bibnamefont {Xiao}},\ and\
  \bibinfo {author} {\bibfnamefont {P.}~\bibnamefont {Bogdan}},\ }\bibfield
  {title} {\enquote {\bibinfo {title} {Multiwavelet-based operator learning for
  differential equations},}\ }\href@noop {} {\bibfield  {journal} {\bibinfo
  {journal} {Advances in neural information processing systems}\ }\textbf
  {\bibinfo {volume} {34}},\ \bibinfo {pages} {24048--24062} (\bibinfo {year}
  {2021})}\BibitemShut {NoStop}%
\bibitem [{\citenamefont {Tripura}\ and\ \citenamefont
  {Chakraborty}(2023)}]{tripura2023wavelet}%
  \BibitemOpen
  \bibfield  {author} {\bibinfo {author} {\bibfnamefont {T.}~\bibnamefont
  {Tripura}}\ and\ \bibinfo {author} {\bibfnamefont {S.}~\bibnamefont
  {Chakraborty}},\ }\bibfield  {title} {\enquote {\bibinfo {title} {Wavelet
  neural operator for solving parametric partial differential equations in
  computational mechanics problems},}\ }\href@noop {} {\bibfield  {journal}
  {\bibinfo  {journal} {Computer Methods in Applied Mechanics and Engineering}\
  }\textbf {\bibinfo {volume} {404}},\ \bibinfo {pages} {115783} (\bibinfo
  {year} {2023})}\BibitemShut {NoStop}%
\bibitem [{\citenamefont {Chen}\ \emph {et~al.}(2023)\citenamefont {Chen},
  \citenamefont {Liu}, \citenamefont {Li}, \citenamefont {Meng},\ and\
  \citenamefont {Chen}}]{chen2023laplace}%
  \BibitemOpen
  \bibfield  {author} {\bibinfo {author} {\bibfnamefont {G.}~\bibnamefont
  {Chen}}, \bibinfo {author} {\bibfnamefont {X.}~\bibnamefont {Liu}}, \bibinfo
  {author} {\bibfnamefont {Y.}~\bibnamefont {Li}}, \bibinfo {author}
  {\bibfnamefont {Q.}~\bibnamefont {Meng}},\ and\ \bibinfo {author}
  {\bibfnamefont {L.}~\bibnamefont {Chen}},\ }\bibfield  {title} {\enquote
  {\bibinfo {title} {Laplace neural operator for complex geometries},}\
  }\href@noop {} {\bibfield  {journal} {\bibinfo  {journal} {arXiv preprint
  arXiv:2302.08166}\ } (\bibinfo {year} {2023})}\BibitemShut {NoStop}%
\bibitem [{\citenamefont {Pathak}\ \emph {et~al.}(2022)\citenamefont {Pathak},
  \citenamefont {Subramanian}, \citenamefont {Harrington}, \citenamefont
  {Raja}, \citenamefont {Chattopadhyay}, \citenamefont {Mardani}, \citenamefont
  {Kurth}, \citenamefont {Hall}, \citenamefont {Li}, \citenamefont
  {Azizzadenesheli} \emph {et~al.}}]{FNOWeatherCast}%
  \BibitemOpen
  \bibfield  {author} {\bibinfo {author} {\bibfnamefont {J.}~\bibnamefont
  {Pathak}}, \bibinfo {author} {\bibfnamefont {S.}~\bibnamefont {Subramanian}},
  \bibinfo {author} {\bibfnamefont {P.}~\bibnamefont {Harrington}}, \bibinfo
  {author} {\bibfnamefont {S.}~\bibnamefont {Raja}}, \bibinfo {author}
  {\bibfnamefont {A.}~\bibnamefont {Chattopadhyay}}, \bibinfo {author}
  {\bibfnamefont {M.}~\bibnamefont {Mardani}}, \bibinfo {author} {\bibfnamefont
  {T.}~\bibnamefont {Kurth}}, \bibinfo {author} {\bibfnamefont
  {D.}~\bibnamefont {Hall}}, \bibinfo {author} {\bibfnamefont {Z.}~\bibnamefont
  {Li}}, \bibinfo {author} {\bibfnamefont {K.}~\bibnamefont {Azizzadenesheli}},
  \emph {et~al.},\ }\bibfield  {title} {\enquote {\bibinfo {title}
  {Fourcastnet: A global data-driven high-resolution weather model using
  adaptive fourier neural operators},}\ }\href@noop {} {\bibfield  {journal}
  {\bibinfo  {journal} {arXiv preprint arXiv:2202.11214}\ } (\bibinfo {year}
  {2022})}\BibitemShut {NoStop}%
\bibitem [{\citenamefont {Yu}(2023)}]{Yu2023}%
  \BibitemOpen
  \bibfield  {author} {\bibinfo {author} {\bibfnamefont {R.}~\bibnamefont
  {Yu}},\ }\bibfield  {title} {\enquote {\bibinfo {title} {Deep learning of
  nonlinear flame fronts development due to darrieus–landau instability},}\
  }\href@noop {} {\bibfield  {journal} {\bibinfo  {journal} {APL APL Machine
  Learning}\ }\textbf {\bibinfo {volume} {1}},\ \bibinfo {pages} {026106}
  (\bibinfo {year} {2023})}\BibitemShut {NoStop}%
\bibitem [{\citenamefont {Ghadami}\ and\ \citenamefont
  {Epureanu}(2022)}]{ghadami2022deep}%
  \BibitemOpen
  \bibfield  {author} {\bibinfo {author} {\bibfnamefont {A.}~\bibnamefont
  {Ghadami}}\ and\ \bibinfo {author} {\bibfnamefont {B.~I.}\ \bibnamefont
  {Epureanu}},\ }\bibfield  {title} {\enquote {\bibinfo {title} {Deep learning
  for centre manifold reduction and stability analysis in nonlinear systems},}\
  }\href@noop {} {\bibfield  {journal} {\bibinfo  {journal} {Philosophical
  Transactions of the Royal Society A}\ }\textbf {\bibinfo {volume} {380}},\
  \bibinfo {pages} {20210212} (\bibinfo {year} {2022})}\BibitemShut {NoStop}%
\bibitem [{\citenamefont {Szegedy}\ \emph {et~al.}(2016)\citenamefont
  {Szegedy}, \citenamefont {Vanhoucke}, \citenamefont {Ioffe}, \citenamefont
  {Shlens},\ and\ \citenamefont {Wojna}}]{inception}%
  \BibitemOpen
  \bibfield  {author} {\bibinfo {author} {\bibfnamefont {C.}~\bibnamefont
  {Szegedy}}, \bibinfo {author} {\bibfnamefont {V.}~\bibnamefont {Vanhoucke}},
  \bibinfo {author} {\bibfnamefont {S.}~\bibnamefont {Ioffe}}, \bibinfo
  {author} {\bibfnamefont {J.}~\bibnamefont {Shlens}},\ and\ \bibinfo {author}
  {\bibfnamefont {Z.}~\bibnamefont {Wojna}},\ }\bibfield  {title} {\enquote
  {\bibinfo {title} {Rethinking the inception architecture for computer
  vision},}\ }in\ \href@noop {} {\emph {\bibinfo {booktitle} {Proceedings of
  the IEEE conference on computer vision and pattern recognition}}}\ (\bibinfo
  {year} {2016})\ pp.\ \bibinfo {pages} {2818--2826}\BibitemShut {NoStop}%
\bibitem [{\citenamefont {Michelson}\ and\ \citenamefont
  {Sivashinsky}(1977)}]{michelson1977nonlinear}%
  \BibitemOpen
  \bibfield  {author} {\bibinfo {author} {\bibfnamefont {D.~M.}\ \bibnamefont
  {Michelson}}\ and\ \bibinfo {author} {\bibfnamefont {G.~I.}\ \bibnamefont
  {Sivashinsky}},\ }\bibfield  {title} {\enquote {\bibinfo {title} {Nonlinear
  analysis of hydrodynamic instability in laminar flames—ii. numerical
  experiments},}\ }\href@noop {} {\bibfield  {journal} {\bibinfo  {journal}
  {Acta astronautica}\ }\textbf {\bibinfo {volume} {4}},\ \bibinfo {pages}
  {1207--1221} (\bibinfo {year} {1977})}\BibitemShut {NoStop}%
\bibitem [{\citenamefont {G.I.Sivashinsky}(1977)}]{SivaEq}%
  \BibitemOpen
  \bibfield  {author} {\bibinfo {author} {\bibnamefont {G.I.Sivashinsky}},\
  }\bibfield  {title} {\enquote {\bibinfo {title} {Nonlinear analysis of
  hydrodynamic instability in laminar flames—i. derivation of basic
  equations},}\ }\href@noop {} {\bibfield  {journal} {\bibinfo  {journal} {Acta
  Astronautica}\ }\textbf {\bibinfo {volume} {4}},\ \bibinfo {pages}
  {1177--1206} (\bibinfo {year} {1977})}\BibitemShut {NoStop}%
\bibitem [{\citenamefont {Kuramoto}(1978)}]{kuramoto1978diffusion}%
  \BibitemOpen
  \bibfield  {author} {\bibinfo {author} {\bibfnamefont {Y.}~\bibnamefont
  {Kuramoto}},\ }\bibfield  {title} {\enquote {\bibinfo {title}
  {Diffusion-induced chaos in reaction systems},}\ }\href@noop {} {\bibfield
  {journal} {\bibinfo  {journal} {Progress of Theoretical Physics Supplement}\
  }\textbf {\bibinfo {volume} {64}},\ \bibinfo {pages} {346--367} (\bibinfo
  {year} {1978})}\BibitemShut {NoStop}%
\bibitem [{\citenamefont {Darrieus}(1938)}]{DARRIEUS1938UNPB}%
  \BibitemOpen
  \bibfield  {author} {\bibinfo {author} {\bibfnamefont {G.}~\bibnamefont
  {Darrieus}},\ }\bibfield  {title} {\enquote {\bibinfo {title} {Propagation
  d’un front de flamme},}\ }\href@noop {} {\bibfield  {journal} {\bibinfo
  {journal} {Unpublished work presented at La Technique Moderne}\ } (\bibinfo
  {year} {1938})}\BibitemShut {NoStop}%
\bibitem [{\citenamefont {Landau}(1988)}]{landau1988theory}%
  \BibitemOpen
  \bibfield  {author} {\bibinfo {author} {\bibfnamefont {L.}~\bibnamefont
  {Landau}},\ }\bibfield  {title} {\enquote {\bibinfo {title} {On the theory of
  slow combustion},}\ }in\ \href@noop {} {\emph {\bibinfo {booktitle} {Dynamics
  of curved fronts}}}\ (\bibinfo  {publisher} {Elsevier},\ \bibinfo {year}
  {1988})\ pp.\ \bibinfo {pages} {403--411}\BibitemShut {NoStop}%
\bibitem [{\citenamefont {Zeldovich}(1944)}]{zeldovich1944selected}%
  \BibitemOpen
  \bibfield  {author} {\bibinfo {author} {\bibfnamefont {Y.}~\bibnamefont
  {Zeldovich}},\ }\bibfield  {title} {\enquote {\bibinfo {title} {Theory of
  combustion and detonation of gases},}\ }in\ \href@noop {} {\emph {\bibinfo
  {booktitle} {Selected Works of Yakov Borisovich Zeldovich, Volume I: Chemical
  Physics and Hydrodynamics}}}\ (\bibinfo  {publisher} {Princeton University
  Press},\ \bibinfo {year} {1944})\BibitemShut {NoStop}%
\bibitem [{\citenamefont {Sivashinsky}(1977)}]{sivashinsky1977diffusional}%
  \BibitemOpen
  \bibfield  {author} {\bibinfo {author} {\bibfnamefont {G.}~\bibnamefont
  {Sivashinsky}},\ }\bibfield  {title} {\enquote {\bibinfo {title}
  {Diffusional-thermal theory of cellular flames},}\ }\href@noop {} {\bibfield
  {journal} {\bibinfo  {journal} {Combustion Science and Technology}\ }\textbf
  {\bibinfo {volume} {15}},\ \bibinfo {pages} {137--145} (\bibinfo {year}
  {1977})}\BibitemShut {NoStop}%
\bibitem [{\citenamefont {Yu}, \citenamefont {Bai},\ and\ \citenamefont
  {Bychkov}(2015)}]{YBB15PRE}%
  \BibitemOpen
  \bibfield  {author} {\bibinfo {author} {\bibfnamefont {R.}~\bibnamefont
  {Yu}}, \bibinfo {author} {\bibfnamefont {X.~S.}\ \bibnamefont {Bai}},\ and\
  \bibinfo {author} {\bibfnamefont {V.}~\bibnamefont {Bychkov}},\ }\bibfield
  {title} {\enquote {\bibinfo {title} {Fractal flame structure due to the
  hydrodynamic darrieus-landau instability},}\ }\href@noop {} {\bibfield
  {journal} {\bibinfo  {journal} {Phys. Rev. E}\ }\textbf {\bibinfo {volume}
  {92}},\ \bibinfo {pages} {063028} (\bibinfo {year} {2015})}\BibitemShut
  {NoStop}%
\bibitem [{\citenamefont {Thual}, \citenamefont {Frisch},\ and\ \citenamefont
  {Hénon}(1985)}]{Thual_Frisch_Henon_poledecomp}%
  \BibitemOpen
  \bibfield  {author} {\bibinfo {author} {\bibfnamefont {O.}~\bibnamefont
  {Thual}}, \bibinfo {author} {\bibfnamefont {U.}~\bibnamefont {Frisch}},\ and\
  \bibinfo {author} {\bibfnamefont {M.}~\bibnamefont {Hénon}},\ }\bibfield
  {title} {\enquote {\bibinfo {title} {Application of pole decomposition to an
  equation governing the dynamics of wrinkled flame fronts},}\ }\href@noop {}
  {\bibfield  {journal} {\bibinfo  {journal} {Journal de Physique}\ }\textbf
  {\bibinfo {volume} {46}},\ \bibinfo {pages} {1485--1494} (\bibinfo {year}
  {1985})}\BibitemShut {NoStop}%
\bibitem [{\citenamefont {Vaynblat}\ and\ \citenamefont
  {Matalon}(2000{\natexlab{a}})}]{Vaynblat_matalon_polestability1}%
  \BibitemOpen
  \bibfield  {author} {\bibinfo {author} {\bibfnamefont {D.}~\bibnamefont
  {Vaynblat}}\ and\ \bibinfo {author} {\bibfnamefont {M.}~\bibnamefont
  {Matalon}},\ }\bibfield  {title} {\enquote {\bibinfo {title} {Stability of
  pole solutions for planar propagating flames: I. exact eigenvalues and
  eigenfunctions},}\ }\href@noop {} {\bibfield  {journal} {\bibinfo  {journal}
  {SIAM J. Appl. Math.}\ }\textbf {\bibinfo {volume} {60}},\ \bibinfo {pages}
  {679--702} (\bibinfo {year} {2000}{\natexlab{a}})}\BibitemShut {NoStop}%
\bibitem [{\citenamefont {Vaynblat}\ and\ \citenamefont
  {Matalon}(2000{\natexlab{b}})}]{Vaynblat_matalon_polestability2}%
  \BibitemOpen
  \bibfield  {author} {\bibinfo {author} {\bibfnamefont {D.}~\bibnamefont
  {Vaynblat}}\ and\ \bibinfo {author} {\bibfnamefont {M.}~\bibnamefont
  {Matalon}},\ }\bibfield  {title} {\enquote {\bibinfo {title} {Stability of
  pole solutions for planar propagating flames: Ii. properties of
  eigenvalues/eigenfunctions and implications to stability},}\ }\href@noop {}
  {\bibfield  {journal} {\bibinfo  {journal} {SIAM J. Appl. Math.}\ }\textbf
  {\bibinfo {volume} {60}},\ \bibinfo {pages} {703--728} (\bibinfo {year}
  {2000}{\natexlab{b}})}\BibitemShut {NoStop}%
\bibitem [{\citenamefont {Olami}\ \emph {et~al.}(1997)\citenamefont {Olami},
  \citenamefont {Galanti}, \citenamefont {Kupervasser},\ and\ \citenamefont
  {Procaccia}}]{Olami_noise}%
  \BibitemOpen
  \bibfield  {author} {\bibinfo {author} {\bibfnamefont {Z.}~\bibnamefont
  {Olami}}, \bibinfo {author} {\bibfnamefont {B.}~\bibnamefont {Galanti}},
  \bibinfo {author} {\bibfnamefont {O.}~\bibnamefont {Kupervasser}},\ and\
  \bibinfo {author} {\bibfnamefont {I.}~\bibnamefont {Procaccia}},\ }\bibfield
  {title} {\enquote {\bibinfo {title} {Random noise and pole dynamics in
  unstable front propagation},}\ }\href@noop {} {\bibfield  {journal} {\bibinfo
   {journal} {Physical Review E}\ }\textbf {\bibinfo {volume} {55}},\ \bibinfo
  {pages} {2649} (\bibinfo {year} {1997})}\BibitemShut {NoStop}%
\bibitem [{\citenamefont {Denet}(2006)}]{denet2006stationary}%
  \BibitemOpen
  \bibfield  {author} {\bibinfo {author} {\bibfnamefont {B.}~\bibnamefont
  {Denet}},\ }\bibfield  {title} {\enquote {\bibinfo {title} {Stationary
  solutions and neumann boundary conditions in the sivashinsky equation},}\
  }\href@noop {} {\bibfield  {journal} {\bibinfo  {journal} {Physical Review
  E}\ }\textbf {\bibinfo {volume} {74}},\ \bibinfo {pages} {036303} (\bibinfo
  {year} {2006})}\BibitemShut {NoStop}%
\bibitem [{\citenamefont {Kupervasser}(2015)}]{Kupervasser_pole_book}%
  \BibitemOpen
  \bibfield  {author} {\bibinfo {author} {\bibfnamefont {O.}~\bibnamefont
  {Kupervasser}},\ }\href@noop {} {\emph {\bibinfo {title} {{Pole Solutions for
  Flame Front Propagation}}}}\ (\bibinfo  {publisher} {Springer International
  Publishing},\ \bibinfo {year} {2015})\BibitemShut {NoStop}%
\bibitem [{\citenamefont {Karlin}(2002)}]{Karlin2002cellular}%
  \BibitemOpen
  \bibfield  {author} {\bibinfo {author} {\bibfnamefont {V.}~\bibnamefont
  {Karlin}},\ }\bibfield  {title} {\enquote {\bibinfo {title} {Cellular flames
  may exhibit a non-modal transient instability},}\ }\href@noop {} {\bibfield
  {journal} {\bibinfo  {journal} {Proceedings of the Combustion Institute}\
  }\textbf {\bibinfo {volume} {29}},\ \bibinfo {pages} {1537--1542} (\bibinfo
  {year} {2002})}\BibitemShut {NoStop}%
\bibitem [{\citenamefont {Creta}\ \emph {et~al.}(2020)\citenamefont {Creta},
  \citenamefont {Lapenna}, \citenamefont {Lamioni}, \citenamefont {Fogla},\
  and\ \citenamefont {Matalon}}]{Creta2020propagation}%
  \BibitemOpen
  \bibfield  {author} {\bibinfo {author} {\bibfnamefont {F.}~\bibnamefont
  {Creta}}, \bibinfo {author} {\bibfnamefont {P.~E.}\ \bibnamefont {Lapenna}},
  \bibinfo {author} {\bibfnamefont {R.}~\bibnamefont {Lamioni}}, \bibinfo
  {author} {\bibfnamefont {N.}~\bibnamefont {Fogla}},\ and\ \bibinfo {author}
  {\bibfnamefont {M.}~\bibnamefont {Matalon}},\ }\bibfield  {title} {\enquote
  {\bibinfo {title} {Propagation of premixed flames in the presence of
  darrieus--landau and thermal diffusive instabilities},}\ }\href@noop {}
  {\bibfield  {journal} {\bibinfo  {journal} {Combustion and Flame}\ }\textbf
  {\bibinfo {volume} {216}},\ \bibinfo {pages} {256--270} (\bibinfo {year}
  {2020})}\BibitemShut {NoStop}%
\bibitem [{\citenamefont {Creta}, \citenamefont {Fogla},\ and\ \citenamefont
  {Matalon}(2011)}]{CRETA2011INST}%
  \BibitemOpen
  \bibfield  {author} {\bibinfo {author} {\bibfnamefont {F.}~\bibnamefont
  {Creta}}, \bibinfo {author} {\bibfnamefont {N.}~\bibnamefont {Fogla}},\ and\
  \bibinfo {author} {\bibfnamefont {M.}~\bibnamefont {Matalon}},\ }\bibfield
  {title} {\enquote {\bibinfo {title} {Turbulent propagation of premixed flames
  in the presence of darrieus–landau instability},}\ }\href@noop {}
  {\bibfield  {journal} {\bibinfo  {journal} {Combustion Theory and Modelling}\
  }\textbf {\bibinfo {volume} {15}},\ \bibinfo {pages} {267--298} (\bibinfo
  {year} {2011})}\BibitemShut {NoStop}%
\bibitem [{\citenamefont {Yu}, \citenamefont {Yu},\ and\ \citenamefont
  {Bai}(2012)}]{YYB12}%
  \BibitemOpen
  \bibfield  {author} {\bibinfo {author} {\bibfnamefont {R.}~\bibnamefont
  {Yu}}, \bibinfo {author} {\bibfnamefont {J.}~\bibnamefont {Yu}},\ and\
  \bibinfo {author} {\bibfnamefont {X.}~\bibnamefont {Bai}},\ }\bibfield
  {title} {\enquote {\bibinfo {title} {{An improved high-order scheme for DNS
  of low Mach number turbulent reacting flows based on stiff chemistry
  solver}},}\ }\href@noop {} {\bibfield  {journal} {\bibinfo  {journal} {J.
  Comput. Phys.}\ }\textbf {\bibinfo {volume} {231}},\ \bibinfo {pages} {5504}
  (\bibinfo {year} {2012})}\BibitemShut {NoStop}%
\bibitem [{\citenamefont {Yu}\ and\ \citenamefont {Bai}(2014)}]{YB14}%
  \BibitemOpen
  \bibfield  {author} {\bibinfo {author} {\bibfnamefont {R.}~\bibnamefont
  {Yu}}\ and\ \bibinfo {author} {\bibfnamefont {X.~S.}\ \bibnamefont {Bai}},\
  }\bibfield  {title} {\enquote {\bibinfo {title} {{A fully divergence-free
  method for generation of inhomogeneous and anisotropic turbulence with large
  spatial variation}},}\ }\href@noop {} {\bibfield  {journal} {\bibinfo
  {journal} {J. Comput. Phys.}\ }\textbf {\bibinfo {volume} {256}},\ \bibinfo
  {pages} {234} (\bibinfo {year} {2014})}\BibitemShut {NoStop}%
\bibitem [{\citenamefont {Yu}\ and\ \citenamefont {Bai}(2013)}]{YB13Numeric}%
  \BibitemOpen
  \bibfield  {author} {\bibinfo {author} {\bibfnamefont {R.}~\bibnamefont
  {Yu}}\ and\ \bibinfo {author} {\bibfnamefont {X.~S.}\ \bibnamefont {Bai}},\
  }\bibfield  {title} {\enquote {\bibinfo {title} {{A semi-implicit scheme for
  large Eddy simulation of piston engine flow and combustion}},}\ }\href@noop
  {} {\bibfield  {journal} {\bibinfo  {journal} {Int. J. Numer. Methods
  Fluids}\ }\textbf {\bibinfo {volume} {71}},\ \bibinfo {pages} {13} (\bibinfo
  {year} {2013})}\BibitemShut {NoStop}%
\bibitem [{\citenamefont {Yu}\ \emph {et~al.}(2017)\citenamefont {Yu},
  \citenamefont {Yu}, \citenamefont {Bai}, \citenamefont {Sun},\ and\
  \citenamefont {Tan}}]{YU2017nonlinear}%
  \BibitemOpen
  \bibfield  {author} {\bibinfo {author} {\bibfnamefont {J.}~\bibnamefont
  {Yu}}, \bibinfo {author} {\bibfnamefont {R.}~\bibnamefont {Yu}}, \bibinfo
  {author} {\bibfnamefont {X.}~\bibnamefont {Bai}}, \bibinfo {author}
  {\bibfnamefont {M.}~\bibnamefont {Sun}},\ and\ \bibinfo {author}
  {\bibfnamefont {J.-G.}\ \bibnamefont {Tan}},\ }\bibfield  {title} {\enquote
  {\bibinfo {title} {Nonlinear evolution of 2d cellular lean hydrogen/air
  premixed flames with varying initial perturbations in the elevated pressure
  environment},}\ }\href@noop {} {\bibfield  {journal} {\bibinfo  {journal}
  {International Journal of Hydrogen Energy}\ }\textbf {\bibinfo {volume}
  {42}},\ \bibinfo {pages} {3790--3803} (\bibinfo {year} {2017})}\BibitemShut
  {NoStop}%
\bibitem [{\citenamefont {Yu}\ and\ \citenamefont
  {Lipatnikov}(2019)}]{YL2019_Equation}%
  \BibitemOpen
  \bibfield  {author} {\bibinfo {author} {\bibfnamefont {R.}~\bibnamefont
  {Yu}}\ and\ \bibinfo {author} {\bibfnamefont {A.~N.}\ \bibnamefont
  {Lipatnikov}},\ }\bibfield  {title} {\enquote {\bibinfo {title}
  {{Surface-averaged quantities in turbulent reacting flows and relevant
  evolution equations}},}\ }\href@noop {} {\bibfield  {journal} {\bibinfo
  {journal} {Phys. Rev. E}\ }\textbf {\bibinfo {volume} {100}},\ \bibinfo
  {pages} {013107} (\bibinfo {year} {2019})}\BibitemShut {NoStop}%
\bibitem [{\citenamefont {Yu}\ \emph {et~al.}(2019)\citenamefont {Yu},
  \citenamefont {Nillson}, \citenamefont {Bai},\ and\ \citenamefont
  {Lipatnikov}}]{YNBL2019_CF}%
  \BibitemOpen
  \bibfield  {author} {\bibinfo {author} {\bibfnamefont {R.}~\bibnamefont
  {Yu}}, \bibinfo {author} {\bibfnamefont {T.}~\bibnamefont {Nillson}},
  \bibinfo {author} {\bibfnamefont {X.}~\bibnamefont {Bai}},\ and\ \bibinfo
  {author} {\bibfnamefont {A.~N.}\ \bibnamefont {Lipatnikov}},\ }\bibfield
  {title} {\enquote {\bibinfo {title} {{Evolution of averaged local premixed
  flame thickness in a turbulent flow}},}\ }\href@noop {} {\bibfield  {journal}
  {\bibinfo  {journal} {Combust. Flame}\ }\textbf {\bibinfo {volume} {207}},\
  \bibinfo {pages} {232--249} (\bibinfo {year} {2019})}\BibitemShut {NoStop}%
\bibitem [{\citenamefont {Yu}\ \emph {et~al.}(2021)\citenamefont {Yu},
  \citenamefont {Nilsson}, \citenamefont {Fureby},\ and\ \citenamefont
  {Lipatnikov}}]{YNCL2020_JFM}%
  \BibitemOpen
  \bibfield  {author} {\bibinfo {author} {\bibfnamefont {R.}~\bibnamefont
  {Yu}}, \bibinfo {author} {\bibfnamefont {T.}~\bibnamefont {Nilsson}},
  \bibinfo {author} {\bibfnamefont {C.}~\bibnamefont {Fureby}},\ and\ \bibinfo
  {author} {\bibfnamefont {A.}~\bibnamefont {Lipatnikov}},\ }\bibfield  {title}
  {\enquote {\bibinfo {title} {Evolution equations for the decomposed
  components of displacement speed in a reactive scalar field},}\ }\href@noop
  {} {\bibfield  {journal} {\bibinfo  {journal} {Journal of Fluid Mechanics}\
  }\textbf {\bibinfo {volume} {911}} (\bibinfo {year} {2021})}\BibitemShut
  {NoStop}%
\end{thebibliography}%
\end{document}